\documentclass[lettersize,journal]{IEEEtran}
\ifCLASSOPTIONcompsoc
  \usepackage[nocompress]{cite}
\else
  \usepackage{cite}
\fi
\ifCLASSINFOpdf
\else
\fi

\usepackage{array}
\usepackage{multirow}
\usepackage{graphicx}
\usepackage{booktabs}
\usepackage{amsmath}
\usepackage[caption=false]{subfig}
\usepackage[autostyle]{csquotes}

\newcommand{\RNum}[1]{\uppercase\expandafter{\romannumeral #1\relax}}

\begin{document}
%
\title{Attribute Annotation and Bias Evaluation in Visual Datasets for Autonomous Driving}
%
%
%
%

\author{David~Fernández~Llorca, Pedro~Frau, Ignacio~Parra, Rubén~Izquierdo and~Emilia~Gómez,
\IEEEcompsocitemizethanks{\IEEEcompsocthanksitem D. Fernández Llorca and E. Gómez are with the Joint Research Centre, European Commission, Sevilla, Spain \protect\\ 
E-mail: david.fernandez-llorca@ec.europa.eu.
\IEEEcompsocthanksitem P. Frau is with Lambda Developments S. L., Madrid, Spain. 
\IEEEcompsocthanksitem D. Fernández Llorca, I. Parra and R. Izquierdo are with the Computer Engineering Department, University of Alcalá, Alcalá de Henares, Madrid, Spain. }
}

%
%

\markboth{Journal of \LaTeX\ Class Files,~Vol.~vv, No.~nn, December~2023}%
{Fernández Llorca \MakeLowercase{\textit{et al.}}: Biases in Visual Datasets for Autonomous Driving}
%



\IEEEtitleabstractindextext{%
\begin{abstract}
This paper addresses the often overlooked issue of fairness in the autonomous driving domain, particularly in vision-based perception and prediction systems, which play a pivotal role in the overall functioning of Autonomous Vehicles (AVs). We focus our analysis on biases present in some of the most commonly used visual datasets for training person and vehicle detection systems. We introduce an annotation methodology and a specialised annotation tool, both designed to annotate protected attributes of agents in visual datasets. We validate our methodology through an inter-rater agreement analysis and provide the distribution of attributes across all datasets. These include annotations for the attributes age, sex, skin tone, group, and means of transport for more than 90K people, as well as vehicle type, colour, and car type for over 50K vehicles. Generally, diversity is very low for most attributes, with some groups, such as children, wheelchair users, or personal mobility vehicle users, being extremely underrepresented in the analysed datasets. The study contributes significantly to efforts to consider fairness in the evaluation of perception and prediction systems for AVs. 

This paper follows reproducibility principles. The annotation tool, scripts and the annotated attributes can be accessed publicly at https://github.com/ec-jrc/humaint\_annotator.
\end{abstract}

}

\maketitle

\IEEEdisplaynontitleabstractindextext

\IEEEpeerreviewmaketitle

\section{Introduction}
\label{sec:introduction}

Autonomous Driving Systems (ADS) rely on accurate detection of persons and vehicles, as well as prediction of their future actions and motions, in order to operate safely and efficiently. Indeed, perception and prediction (a.k.a. dynamic scene understanding) is one of the key operational layers of Autonomous Vehicles (AVs) \cite{Llorca2023}. This task primarily utilizes information obtained from digital cameras and range sensors, such as LiDAR or radar. Generally, the most sophisticated perception and prediction systems are built upon deep learning models, which significantly depend on datasets for their development and evaluation. Although these approaches are becoming increasingly advanced and effective, current design and evaluation methods overlook the ethical principle of \emph{fairness}. In the machine learning domain, this principle is generally interpreted as \emph{equality of opportunity} \cite{Hardt2016}, or \emph{unfair bias} \cite{Songul2019}, and it refers to the potential variation of performance levels across different population demographics \cite{Llorca2021}.

The risk of unequal performance can be attributed to two main components. Firstly, the bias in datasets. The performance of machine learning models is heavily influenced by the number of training samples. Therefore, categories that are underrepresented in datasets will inherently pose a greater challenge for the model to learn \cite{liu2019largescale}, \cite{Corrales2021}. Secondly, the lack of algorithmic fairness that aim to counteract the bias in the datasets. Algorithmic solutions encompass various strategies, including different pre-process, in-process and post-process mechanisms \cite{Pessach2020}. However, as stated in \cite{Yang2020} algorithmic interventions alone are unlikely to be the most effective path towards fairness in machine learning, making dataset interventions necessary. Furthermore, most algorithmic approaches are supervised and require explicit annotation of protected attributes, which further underscores the need for intervention at the dataset level \cite{Yang2020}.

Key attributes for studying bias in visual datasets \footnote{While the potential bias in datasets from other data modalities, such as range-based (e.g., LiDAR, radar) or infrared spectrum, may also be significant, the analysis of these ones is beyond the scope of this paper.} for autonomous driving may include sex/gender, age, or skin tone for pedestrians, and colour, type, or size for vehicles. For example, a dataset that primarily features lighter-skinned people \cite{Wilson2019} or certain age groups \cite{Brandao2019} may not accurately reflect the diversity of people on the road in other scenarios. Similarly, a dataset that primarily features small size vehicles of certain colours may not accurately represent the range of vehicles encountered in other real-world situations. The consequences of dataset bias in the field of autonomous driving can be significant. Inaccurate detection of certain type of pedestrians or vehicles, as well as imprecise motion prediction, can result in varying behaviours of autonomous vehicles towards different road agents \cite{Llorca2021}. This could potentially lead to disparities in accident and injury rates, thereby producing an unfair outcome. 

Despite its importance, the study of bias in the context of perception for AVs has been relatively neglected in the literature, with some exceptions \cite{Wilson2019}, \cite{Brandao2019}, \cite{Li2023}. This is not the case in the computer vision community, where there is a considerable body of work focused on addressing the issue of underrepresentation of certain demographic groups in datasets \cite{Fabbrizzi2022}. 

In the general field of machine learning, proposed standards for model \cite{Mitchell2019} and data \cite{gebru2021datasheets} transparency reporting advocate the disclosure of performance metrics, disaggregated by relevant population groups. Furthermore, numerous strategies to mitigate bias require labelling attributes in at least a portion of the training dataset \cite{Schumann2021}. To categorize different demographic groups, it is necessary to have a certain level of granularity in the attributes of the agents. In terms of fairness, some attributes may be protected, that is, the performance of the model should be agnostic with respect to them \cite{Hardt2016}.

Attribute labelling is thus a crucial step in identifying biases in visual datasets. The annotation of attributes such as age, sex/gender, and skin tone in person datasets, or car model in vehicle datasets, can pose significant challenges. These challenges stem from the subjective perception of these characteristics, potential biases of the annotators (including unconscious biases), and the ethical and privacy considerations involved in the process. Therefore, the annotation procedure plays a pivotal role, especially when labelling large amounts of data that require multiple annotators working in parallel on non-overlapping datasets. 

The main contributions of this paper are as follows. Firstly, we present a new set of annotations for subsets of some of the most commonly utilised visual datasets for training perception and prediction systems for different road agents in the autonomous driving domain. We focus on both person and vehicle datasets that meet an inclusion criterion, taking into account minimum image resolution and diversity. Attributes of over 90K persons and more than 50K vehicles have been annotated, including previously unconsidered  attributes such as the individual's mean of transport or group assignment. Secondly, we introduce a specialised annotation tool that allows multiple users to simultaneously annotate additional attributes of road agents in visual datasets for AVs, including the proposal for a common standardised file format for the recorded data and associated attributes. Additionally, we have developed an annotation methodology designed to minimize common errors, biases, and discrepancies among annotators. Finally, we validate our annotation methodology by analysing inter-rater agreement, and we present the final distribution of attributes across all datasets and identify the most important biases. The annotation tool, scripts and the annotated attributes are publicly available. The presented work serves as a significant step towards future work in addressing the fairness of perception systems for autonomous vehicles.

\section{Related Work}\label{sec:sota}
In this section, we review the principal studies related to attribute annotation and bias analysis in datasets within the broader context of computer vision, and specifically in the case of autonomous driving. While this topic has also been explored in other non-visual domains \cite{Fabris2022}, such analysis falls outside the scope of this paper as we focus on the annotation of visual attributes. 

\subsection{Computer Vision}
The problem of dataset bias has been widely addressed in the computer vision context in multiple application domains \cite{Fabbrizzi2022}. For example, in face recognition and gender classification \cite{Buolamwini2018}, face attribute detection \cite{JungRyu2018}, facial expression and emotion recognition \cite{Rhue2018}, \cite{Xu2020}, human activity recognition \cite{choi2019sdn}, fine-grained vehicle type classification \cite{Corrales2021}, or image captioning \cite{Hendricks2018}, among others. 

Similarly, numerous studies focus on bias-aware annotation of protected attributes in new or pre-existing datasets. In \cite{Buolamwini2018}, the authors released a new balanced dataset, the Pilot Parliaments Benchmark (PPB,) by collecting and annotating data related to gender and the skin tone from photos of members from six different parliaments, with a total of 1.2K images. They also annotated these attributes in two existing datasets, IJB-A and Adience. 
A subset of about 108K images, extracted from the YFCC-100M dataset, was utilised in \cite{Karkkainen2021} to label race, gender and age attributes for face recognition systems. They used crowd workers for the annotation process, with three annotators per image. Consensus among annotator determined the ground truth. Disagreements led to reassignment or discarding of images. A model trained on these initial annotations was used to refine them, with manual re-verification for any discrepancies. As stated in \cite{Fabbrizzi2022} this process does not guarantee correct annotations. The same YFCC-100M dataset was used in \cite{Merler2019} to create a subset of almost 1M images including age, gender, skin tone, a set of craniofacial ratios, pose and illumination as protected attributes. However, they used automatic detection tools for the annotation process, which incorporated the biases of the datasets used to train these tools. Similar attributes were manually annotated for a smaller number of images (approximately 40K) in \cite{Georgopoulos2020}, also in the context of face analysis.

In \cite{Yang2020}, the authors employed crowdsourcing to manually annotate gender, skin colour and age attributes from the \enquote{person} subtree of the ImageNet dataset. Each image was annotated by at least two workers and consensus was necessary to accept the annotations. They also implemented a quality control process to exclude workers with high error rates. Similarly, in \cite{Schumann2021} a randomly selected subset of images from the Open Images dataset of about 100K images was annotated for the ‘person’ class, with a focus on attributes of gender and age. Specifically, the categories of person, man, woman, boy and girl were selected. The annotators were instructed about the different labels, including instances where the 'unknown' label might be the appropriate choice. In \cite{Hazirbas2022}, a new dataset was introduced, comprising videos from over 3K subjects. The subjects self-annotated the attributes of gender and age, while a group of annotators labelled the skin tone. 

Finally, it is worth noting that although there are specific tools for the analysis and detection of automatic biases in visual datasets, such as Amazon SageMaker Clarify \cite{SageMaker}, Google's Know Your Data \cite{Google}, or more recently, the REVISE tool \cite{REVISE2022}, most of the aforementioned works did not specify the particular annotation tool used for the dataset bias identification process. The use of specific ad-hoc tools for each case is the most common approach. Moreover, most studies reviewed have somewhat neglected a thorough analysis of annotator reliability, such as inter-rater agreement, particularly when multiple annotators label the same or distinct data.

\subsection{Autonomous Driving}
In the field of autonomous driving, there is a limited body of work focusing on fairness or bias analysis. In \cite{Brandao2019}, the author augmented the annotations of the INRIA Person Dataset by including gender (male or female) and age (child or adult) categories. The findings highlighted a significant underrepresentation of the child category compared to adults, and a lesser number of female agents compared to male ones.  

In \cite{Wilson2019}, the authors annotated pedestrians in the BDD100K dataset according to two different skin tones (light or dark). They used bounding boxes of a minimal size and collected the annotations using crowd workers for the training set ($\sim5000$ pedestrians), while using their own annotations for the validation set ($\sim500$ pedestrians). For the crowdsourced annotations, they only took into account those agents where a consensus was reached. They showed that the annotated subset of the BDD100K dataset contained more than three times as many individuals with light skin compared to those with dark skin. Recently, the same dataset was utilised in \cite{REVISE2022} to conduct a geography-based bias analysis, which was correlated to income and weather conditions.  

Finally, we highlight a recent study \cite{Li2023} that focused on gender, age and skin tone as protected attributes in the following pedestrians datasets: CityPersons, Eurocity Persons (Day and Night) and BDD100k. Two annotators independently labelled gender (male or female) and age (adult or child) for the same agents with certain bounding box size. They annotated approximately 16K and 20K labels corresponding to gender and age, respectively, and use the available annotations for skin tone from the BDD100k dataset \cite{Wilson2019}. They found a slight difference in the representation of males compared with to females, and a significant bias towards adults compared to to children across all datasets. 

\section{Datasets}\label{sec:datasets}
Current state-of-the-art scene understanding or visual perception systems for autonomous vehicles are learning-based or data-driven, and rely heavily on the availability of public datasets. In addition, these datasets serve as the basis for various benchmarking processes that are very useful for the academic and industrial community. Unlike previous works that have focused the bias analysis only on pedestrian detection, and using a single dataset (i.e., the INRIA Person Dataset in \cite{Brandao2019} and the BDD100K in \cite{Wilson2019}), we propose to broaden the scope by also considering the detection of cyclists and vehicles, and including multiple datasets. Thus, our goal is more ambitious, as we aim to develop a tool and validate a methodology that can be applied holistically to annotate additional attributes in any visual dataset for the autonomous driving domain. With this approach, we also obtain more general bias analysis results.

We have selected up to six different datasets that are highly representative of the current computer vision data ecosystem in the field of autonomous driving. Our inclusion criteria for the selection of the datasets take into account several parameters, such as minimum image resolution, and diversity in geographical distribution and type of agents, including datasets from Europe, the US and China, and agents such as pedestrians, riders, users of personal mobility devices, wheelchairs and different types of vehicles. 

\begin{figure}[t]
\centering
\includegraphics[width=0.49\textwidth]{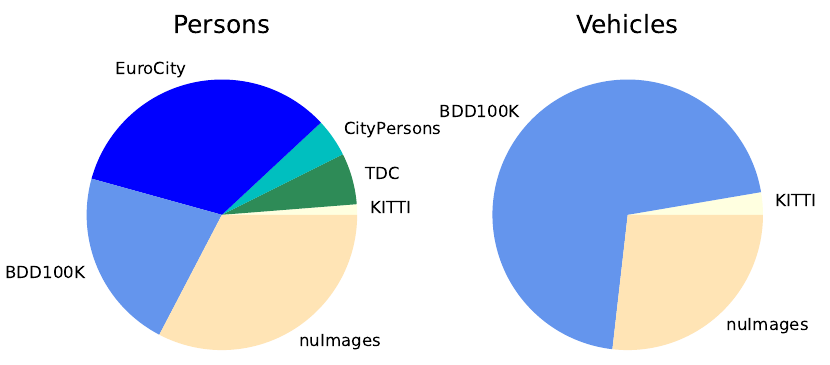}
\caption{Original distribution of annotated samples for the persons and vehicles categories.}
\label{fig:pie1}
\end{figure}

It is important to note that most datasets were published as a challenge for researchers with an associated benchmarking process. In such cases, annotations (i.e., bounding boxes) are only available for training and validation datasets, so attributes cannot be annotated for test datasets. The six datasets that have been selected for the annotation and bias study process are briefly described below. An overall but detailed comparison is presented in Table \ref{tab:datasets}, and the general distributed is depicted in Fig \ref{fig:pie1}.  

\begin{table*}[!ht]
\caption{Comparison of person and vehicle annotated datasets in the autonomous driving domain}
\label{tab:datasets}
\centering
\begin{tabular}{ccccccc}
\toprule  
    & KITTI \cite{Kitti2012} & TDC \cite{TDC-2016} & CityPersons \cite{CityPersons2017} & EuroCity Persons \cite{Eurocity2019} & BDD100K \cite{bdd100k-2020} & nuImages \cite{nuImages}\\
    \midrule
year & 2012 & 2016 & 2017 & 2019 & 2020 & 2020 \\
annotated sets & tr & tr, val, ts & tr, val & tr, val & tr, val & tr, val\\   resolution & $1240 \times 376$ & $2048 \times 1024$ & $1920 \times 1024$ & $2048 \times 1024$ & $1280 \times 720$ & $1600 \times 900$ \\
lighting conditions & day & day & day & day, night & day, night & day, night\\
weather & dry & dry & dry & dry, wet & dry, wet & dry, wet, snow\\
\# countries & 1 & 1 & 3 & 12 & 1 & 2\\
\# cities / areas & 1 & 1 & 27 & 31 & 4 & 2\\
\# annotated persons & $\sim 6336$ & $\sim 31115$ & $\sim 23089$ & $\sim 171145$ & $\sim 109777$ & $\sim 165587$\\
\# annotated vehicles & $\sim 33261$ & -- & -- & -- & $\sim 875024$ & $\sim 332637$\\
\bottomrule
\end{tabular}
\end{table*}

\subsection{KITTI Dataset}\label{subsec:kitti}
The KITTI Dataset is probably one of the most well-known datasets in the context of autonomous vehicles. Presented in 2012 \cite{Kitti2012}, it provides a suite of vision tasks such as stereo, optical flow, visual odometry, etc., built using an autonomous driving platform. It contains the object detection dataset, including mid-resolution images (1240 $\times$ 376) and bounding boxes for pedestrians and vehicles. It was recorded in the city of Karlsruhe and nearby areas in Germany. The image resolution, number of agents and diversity are very limited. Although it does not reflect the current state of the art, this dataset is by far the most widely used in this domain. For the object detection task, the training set provides the location of about $6336$ agents corresponding to people, including pedestrians, sitting persons and cyclists, and about $33261$ vehicles, including cars, vans trucks, and tram categories. 

\subsection{Tsinghua-Daimler Dataset }\label{subsec:tsinghua}
The Tsinghua-Daimler (TDC) dataset mainly focuses on cyclists and other riders. It was presented in 2016 \cite{TDC-2016}, and recorded using a moving vehicle in the urban traffic of Beijing in China. In this case, the labels are provided for the training, validation and test datasets, including a total of about $31115$ agents corresponding to the person category, mainly including cyclists, but also pedestrians. 

\subsection{CityPersons Dataset }\label{subsec:citypersons}
The CityPersons dataset was presented in 2017 \cite{CityPersons2017} as a new set of person annotations on top of the Cityscapes dataset \cite{Cityscapes2016}. Citiscapes, originally presented in 2016, contained a diverse set of stereo video sequences recorded in 50 different cities. CityPersons annotations correspond to a subset of 27 different cities (most of them from Germany), and it includes a training and validation datasets with about $23089$ bounding boxes and labels for persons, including pedestrians, riders, sitting persons and others (unusual postures). 

\subsection{EuroCity Persons Dataset}\label{subsec:eurocity}
The EuroCity Persons dataset, presented in 2019 \cite{Eurocity2019} was specifically conceived to provide highly diverse annotations of pedestrians, cyclists and other riders in urban traffic scenes, including 31 cities of up to 12 different European countries. It also provides data in night-time conditions. Including daytime and nighttime conditions, the training and validation datasets contain about $171145$ annotations corresponding to persons. 

\subsection{BDD100K Dataset}\label{subsec:bdd100k}
The Berkeley Deep Drive dataset (BDD) was presented in 2020 \cite{bdd100k-2020}. It consists of over 100K video clips recorded from more than 50K rides covering multiple cities and nearby areas in the US (e.g., New York, Oakland, Berkeley, San Francisco, San Jose, etc.). It includes up to six different weather conditions, and three different times of day, including nighttime sequences. For each video, they provide bounding box annotations for one reference frame, including pedestrians and riders for the person category, cars, trucks, buses, trains and motorcycles for the vehicle category, as well as other elements such as traffic lights and signs. In total, the training and validation datasets include about $109777$ and $875024$ samples from the person and vehicle categories respectively.

\subsection{nuImages Dataset}\label{subsec:nuscenes}
The nuTonomy Scenes (nuScenes) dataset was recorded in Boston (US) and Singapore. Presented in 2020 \cite{nuScenes2020}, it was conceived as the first multimodal dataset providing data from a complete autonomous vehicle sensor suite (6 cameras, 5 radars, and 1 lidar). The nuScenes dataset was complemented with nuImages dataset the \cite{nuImages}, which offers a total of 93K 2d annotated images from the 6 cameras, with higher variability. These annotated images encompass conditions such as rain, snow and nighttime, and provide up to 23 different classes of objects. This includes pedestrians and riders within the person category, and cars, trucks, trailers, motorcycles, and buses within the vehicle category. It also offers object attributes, some of them very well aligned with our annotations such as adult, child, personal mobility device or wheelchair. This is a very large-scale dataset that contains a huge number of annotations. The training and validation datasets include about $165587$ of annotated samples for the person category, and about $332637$ for the vehicle one. 


\begin{figure*}[!t]
\centering
\includegraphics[width=0.99\textwidth]{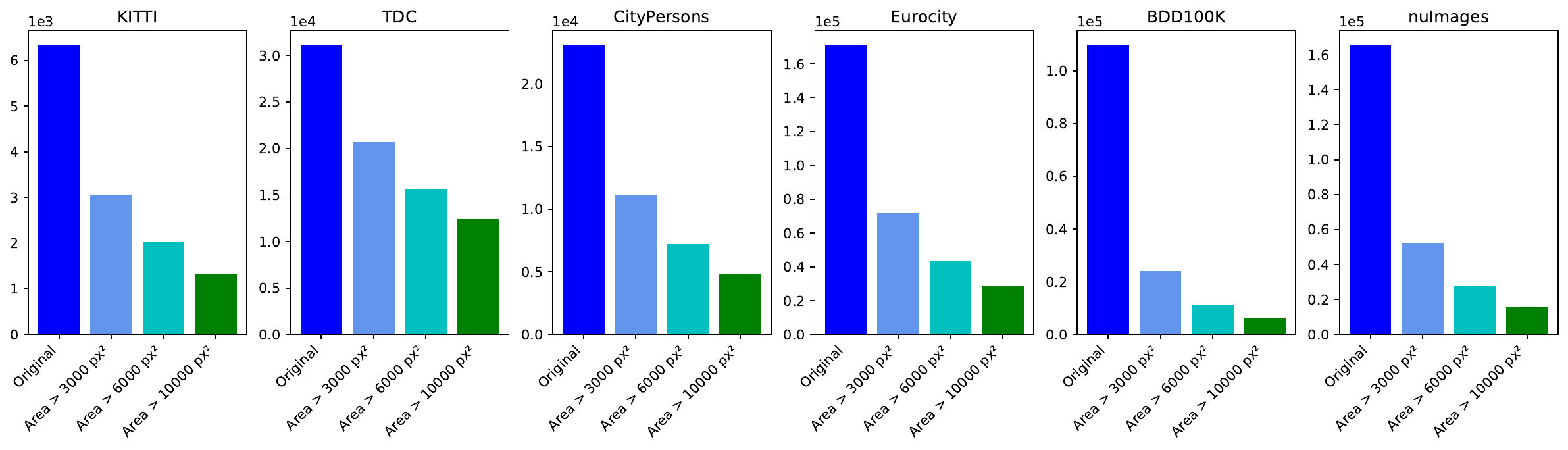}
\caption{Distribution of annotated samples filtered by area for persons datasets.}
\label{fig:barsPers}
\end{figure*}

\begin{figure}[!t]
\centering
\includegraphics[width=0.49\textwidth]{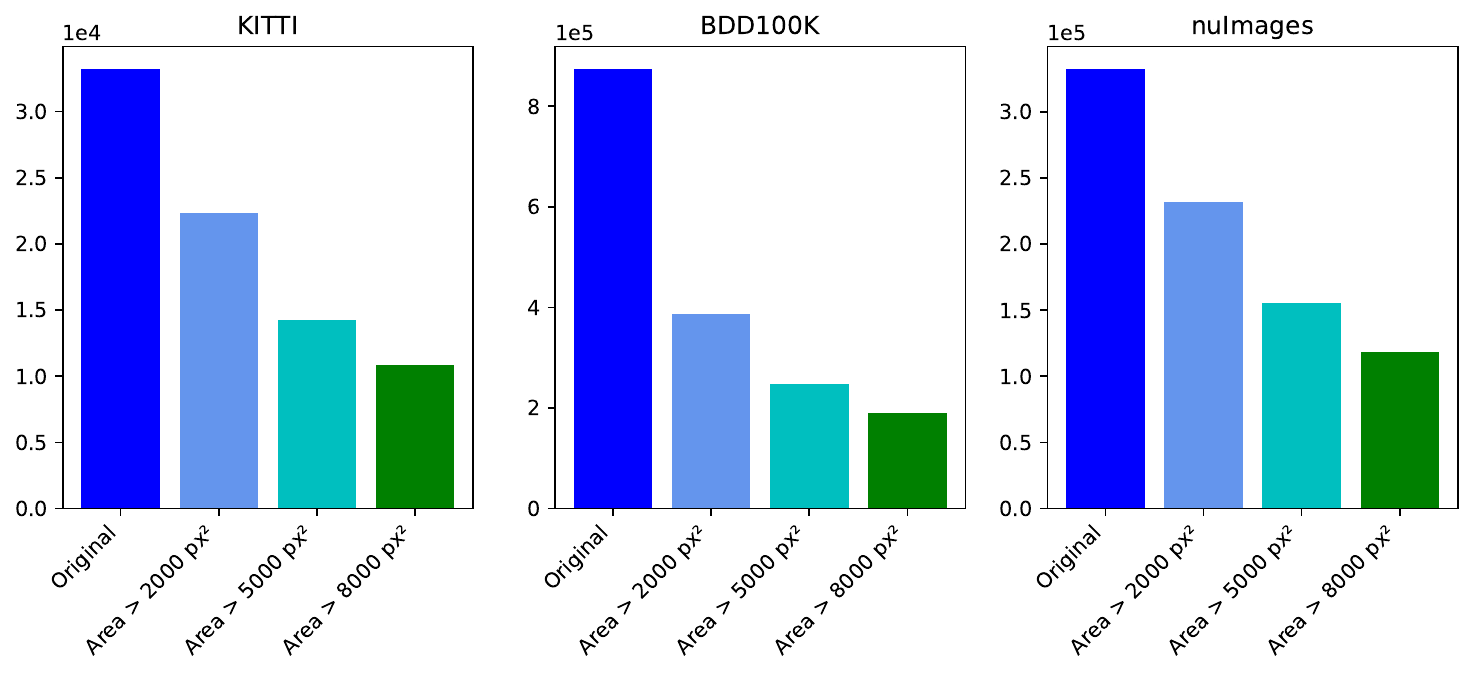}
\caption{Distribution of annotated samples filtered by area for vehicles datasets.}
\label{fig:barsVeh}
\end{figure}

\section{Data Selection}\label{sec:data}
Manual data annotation is a costly and time-consuming process. The manual effort and associated costs required to label images also depend on many factors, such as the complexity of the images, the number and type of agents, the annotation attributes, etc. \cite{Vijayanarasimhan2009}. In this project, taking into account the available resources and the number of attributes required to be annotated per agent, we were able to target an approximate number of $\sim$140K agents plus a minimum of $6\%$ of the samples annotated by at least 5 different annotators to carry out the inter-agreement analysis. One of our hypotheses is that possible biases will be more prevalent for persons. In addition, the number of data sets for persons is larger. Therefore, we  assigned roughly 2/3 of the target samples to the people class ($\sim$90K) and 1/3 to the vehicle class ($\sim$50K).

For attribute labeling to be effective, agents must be visible at a reasonable resolution. There is preliminary evidence that, in the case of small image sizes, the level of disagreement between several annotators in labeling attributes (such as skin tone) can be considerably higher \cite{Wilson2019}. As the number of agents available in all datasets is well above the target of 140K, we first studied the distribution of the samples according to the size of the bounding box area. For the person category we counted the number of agents for bounding box sizes greater than or equal to $3000$, $6000$ and $10000$ pixels. For the vehicle category, our preliminary analysis showed that with smaller sizes it was possible to clearly identify the  attributes under study, so we analyzed those samples with bounding box areas greater than or equal to $2000$, $5000$ and $8000$ pixels. The distribution of annotated samples filtered by area for persons and vehicles datasets are depicted in Figs. \ref{fig:barsPers} and \ref{fig:barsVeh} respectively. As can be seen, the relative number of samples of persons with a resolution higher than $10000$ pixels is particularly low for the BDD100K and nuImages datasets, which can be explained by the lower image resolution used (see Table \ref{tab:datasets}). 

\begin{figure}[!t]
\centering
\includegraphics[width=0.35\textwidth]{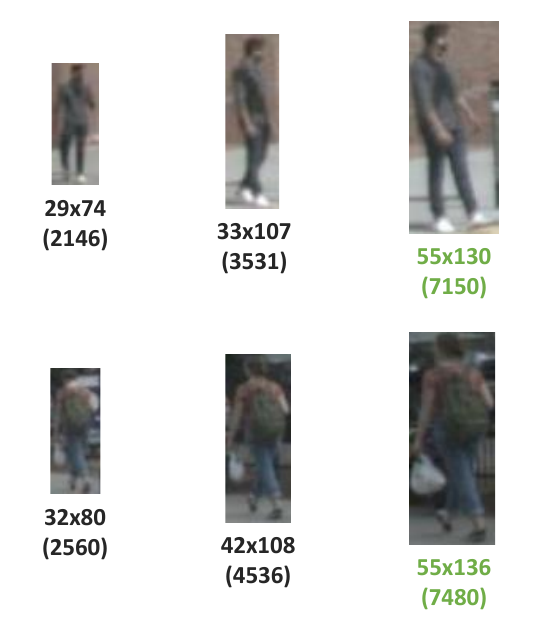}
\caption{Two examples of different sizes of bounding boxes for persons datasets (resolutions and areas). Smaller resolutions do not allow for clear identification of sex or skin tone.}
\label{fig:resPersons}
\end{figure}

\begin{figure}[!t]
\centering
\includegraphics[width=0.4\textwidth]{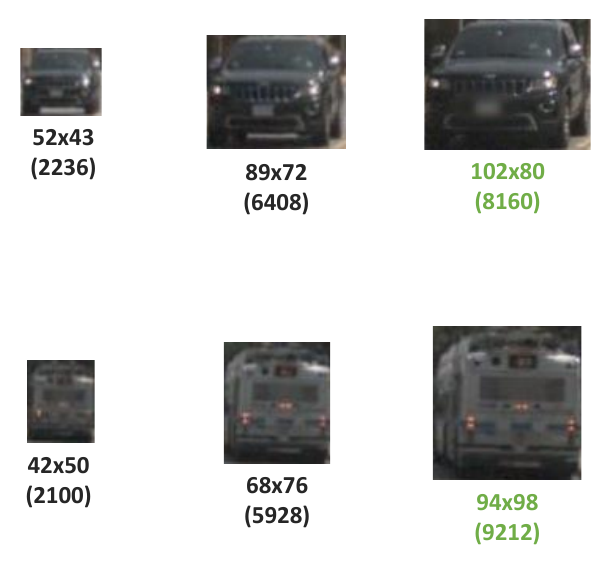}
\caption{Two examples of different sizes of bounding boxes for vehicles datasets (resolutions and areas). Smaller resolutions do not allow for clear identification of vehicle type or colour.}
\label{fig:resVehicles}
\end{figure}

The specific distribution of annotated samples filtered by area for the persons and vehicles datasets is shown in Tables \ref{tab:datapersons} and \ref{tab:datavehicles} respectively, including the tentative final targets, the coverage with respect to the available samples for the selected bounding box area, and the minimum number of samples to be annotated by the annotators to perform the inter-rater reliability analysis. On the one hand, for the vehicles dataset, the number of samples available with the maximum bounding box area ($8000$ pixels) is more than enough to reach a minimum of $50000$ samples. Therefore, we filtered out all samples with an area less than $8000$ pixels (a minimum square bounding box size of about 283 $\times$ 283 pixels). On the other hand, for the persons dataset, the number of samples with an area greater than $10000$ pixels (as in \cite{Wilson2019} for skin tone annotation), is less than $75000$ samples. Therefore, to reach the goal of $90000$ samples, it was  necessary to reduce the filtering threshold to $6000$ pixels (a minimum bounding box size of about 173 $\times$ 346 pixels). In a preliminary study, we found that this size was sufficient to correctly identify the attributes under study, provided that visibility conditions were adequate. With the aforementioned thresholds, we attempted to annotate on average 83\% of the person samples with a bounding box area greater than $6000$ pixels, and 16\% of the vehicle samples with a bounding box area greater than $8000$ pixels. We depict two different examples corresponding to different bounding box sizes for the persons and vehicles datasets in Figs. \ref{fig:resPersons} and \ref{fig:resVehicles} respectively. As can be observed, the smaller resolutions do not allow to clearly differentiate attributes such as sex or skin tone for persons, or type and colour for vehicles.

The large number of samples to be labelled requires the involvement of multiple non-overlapping annotators, labelling different subsets of the datasets (one-way model \cite{tenHove2023}). In total, 5 different annotators have been involved. However, before carrying out the non-overlapping labelling, it is necessary to assess the degree of agreement between the different annotators, and once validated, to continue with the labelling of the remaining samples. For this purpose, and based on the available resources, a minimum of 6\% of the total number of  samples was defined to be labelled by all annotators (two-way model \cite{tenHove2023}). As shown in Tables \ref{tab:datapersons} and \ref{tab:datavehicles}, this amounts to a total of $5400$ labelled samples per annotator for the person databases ($27000$ annotations in total) and a total of $3000$ samples for the vehicle databases ($15000$ annotations in total). These samples will be used to perform the inter-rater reliability analysis.


\newcolumntype{D}{>{\raggedright\arraybackslash}m{0.08\textwidth}}
\newcolumntype{a}{>{\centering\arraybackslash}p{0.08\textwidth}}
\begin{table*}
\caption{Distribution of annotated samples filtered by area for persons datasets, including the final goal and the number of required annotations per annotator.}
\footnotesize
\centering
\scalebox{1.0}{
\begin{tabular}{Daaaaaaa}
\toprule
\textbf{Dataset} & \textbf{Original} & \textbf{Area $>$ 3000 $px^2$} & \textbf{Area $>$ 6000 $px^2$} & \textbf{Area $>$ 10000 $px^2$} &  \textbf{Final Goal} &  \textbf{Coverage$^a$} & \textbf{Inter-rater agreement$^b$} \\
\midrule 
KITTI & 6336 & 3050 & 2023 & 1333 & \textbf{1000} & 49\% & 60 \\
\midrule
TDC & 31115 & 20710 & 15627 & 12444 & \textbf{6000} & 38\% & 360  \\
\midrule
CityPersons & 23089 & 11135 & 7238 & 4822 & \textbf{6000} & 83\% & 360  \\
\midrule
Eurocity & 171145 & 72274 & 43955 & 28516 & \textbf{42000} & 96\% & 2520  \\
\midrule
BDD100K & 109777 &  24036 & 11318 & 6269 & \textbf{10000} & 88\% & 600  \\
\midrule
nuImages & 165587 & 51978 & 27695 & 16177 & \textbf{25000} & 90\% & 1500  \\
\bottomrule \\
Total & 507049 & 183183 & 107856 & 69561 & \textbf{90000} & 83\% & 5400  \\
\bottomrule \\
\multicolumn{7}{l}{\footnotesize{$^a$ Coverage with respect to the available samples with bounding box area $>6000$ $px^{2}$.}} \\
\multicolumn{7}{l}{\footnotesize{$^b$ $6\%$ of the samples annotated by 5 annotators to analyze inter-rater agreement.}} 
\end{tabular}
}
\label{tab:datapersons}   
\end{table*}

\begin{table*}
\caption{Distribution of annotated samples filtered by area for vehicles datasets, including the final goal and the number of required annotations per annotator.}
\footnotesize
\centering
\scalebox{1.0}{
\begin{tabular}{Daaaaaaa}
\toprule
\textbf{Dataset} & \textbf{Original} & \textbf{Area $>$ 2000 $px^2$} & \textbf{Area $>$ 5000 $px^2$} & \textbf{Area $>$ 8000 $px^2$} &  \textbf{Final Goal} &  \textbf{Coverage$^a$} & \textbf{Inter-rater agreement$^b$} \\
\midrule 
KITTI & 33261 & 22315 & 14258 & 10863 & \textbf{2000} & 18\% & 120 \\
\midrule
BBD100K & 875024 & 387183 & 248131 & 190074 & \textbf{30000} & 16\% & 1800  \\
\midrule
nuImages & 332637 & 231468 & 155669 & 118471 & \textbf{18000} & 15\% & 1080  \\
\bottomrule \\
Total & 1240922 & 640966 & 418058 & 319381 & \textbf{50000} & 16\% & 3,000  \\
\bottomrule \\
\multicolumn{7}{l}{\footnotesize{$^a$ Coverage with respect to the available samples with bounding box area $>8000$ $px^{2}$.}} \\
\multicolumn{7}{l}{\footnotesize{$^b$ $6\%$ of the samples annotated by 5 annotators to analyze inter-rater agreement.}} 
\end{tabular}
}
\label{tab:datavehicles}   
\end{table*}

\section{Annotation}\label{sec:annotation}
The description of the annotation tool and methodology goes here.

\subsection{Data Formatting and Annotation Tool}\label{subsec:tool}
For this work, we created a dedicated tool that enables multiple users to annotate additional attributes of agents (i.e., persons and vehicles) in visual datasets designed for autonomous driving. The tool was designed to seamlessly interact with the datasets described in Section \ref{sec:datasets}. The configuration of the tool requires the pre-definition of the final goals, the number of annotators, and the minimum number of samples for inter-rater agreement analysis for each dataset (Tables \ref {tab:datapersons} and \ref {tab:datavehicles}). After these requirements are pre-defined, the annotation process becomes fully transparent to annotators for both inter-agreement and non-overlapping samples. That is, once one of the users has reached the minimum number of inter-agreement samples, the next annotated samples for that user are automatically assigned to the non-overlapping set.

There is currently no standardised format for storing annotation data for these datasets. Each dataset uses its own format for storing image labels and metadata. Therefore, we cannot build a single interpreter to interact with all datasets. To make the data more portable, easy to read and use, and hopefully establish a standard for future datasets, we created a specific parser to convert all datasets annotations data files to a common dictionary organizing the data in \texttt{json} files. The parser is publicly available, and the proposed standardised data format for image metadata, generic agent data and new attributes is detailed in Appendix \ref{appendix:dataformat}. 

The annotation tool was implemented using a backend based on Flask, which mainly serves as a communication interface between the database, the frontend and the storage. The frontend was developed in Javascript as a web application. It includes all visualisation and user interaction functionalities. The entire system was deployed on an internal local server. The tool is publicly available. 

\begin{figure}[!t]
\centering
\includegraphics[width=0.49\textwidth]{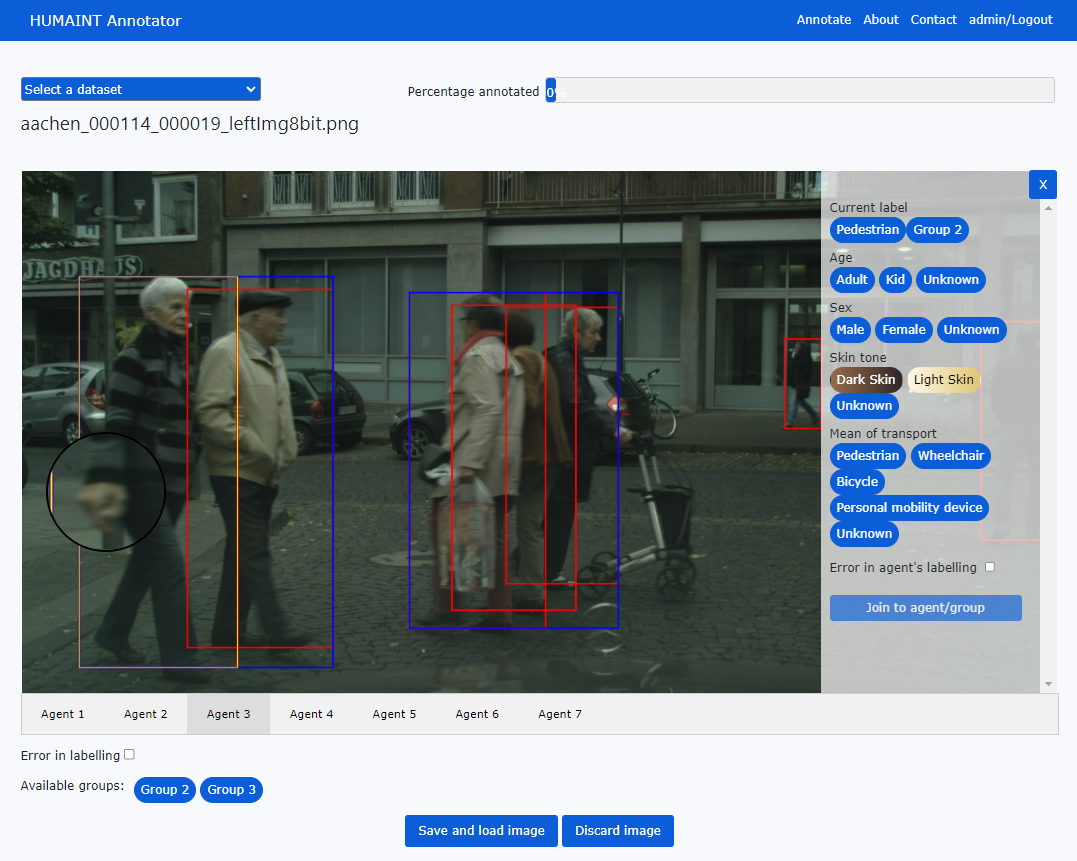}
\caption{Annotation tool interface. Example of a persons dataset. All active agents (minimum size) must be annotated by clicking and selecting the different attributes. In this example agents have been pre-configured in two different groups. The tool is available at: https://github.com/ec-jrc/humaint\_annotator.}
\label{fig:annotationTool}
\end{figure}

Broadly speaking, the annotation procedure implemented in the tool is as follows. Once the user has logged into the annotation web interface and selected a dataset, the tool automatically provides images of the dataset to be annotated. As can be seen in Fig. \ref{fig:annotationTool}, the interface displays the bounding box of each agent that meets the minimum size criterion. By clicking on each agent or using the agent tabs below the image, the user can select the agent and assign different attributes to it from the attributes menu. In addition, a zoom function is available to inspect the image in detail. Once all agents have been annotated, the system will provide a new image, until the final target for each dataset has been met.

Finally, there are two features that have been automated in the tool. First, the pre-assignment of groups of agents, which is based on a similarity criterion according to the position and size of the agent's bounding boxes. This automatic assignment can be manually corrected during the annotation process. Second, the automatic copying of agent attributes into sequence in the nuImages dataset. In this case, the key frame is labelled, and the same attributes are copied to the rest of the frames in the sequence from the unique agent identifier.

\subsection{Attributes}\label{subsec:attributes}
In this work, we focus on the intrinsic attributes of agents that have an impact on vision-based perception and prediction systems. The appearance of the agent is the most prominent feature typically used by detection systems of this type, which learn from samples in datasets collected from real-world driving. The attributes of agents, including their appearance, can have a significant effect on the functioning of the perception system. Therefore, if the datasets used to train the model are not well balanced, the performance of the system may be compromised depending on the type of agent. 

This effect can also occur in systems for predicting the motion and actions of agents. Most predictive perception systems are based on learning an agent behaviour model from data. The behaviour of the agents, including their possible future actions and motions, depends on both intrinsic and extrinsic factors~\cite{Izquierdo22}. While our focus is on intrinsic features that may affect agent behaviour, other extrinsic factors such as scene layout, traffic, lighting, and weather conditions also play a key role.

\begin{figure*}[!th]
\centering
\includegraphics[width=0.9\textwidth]{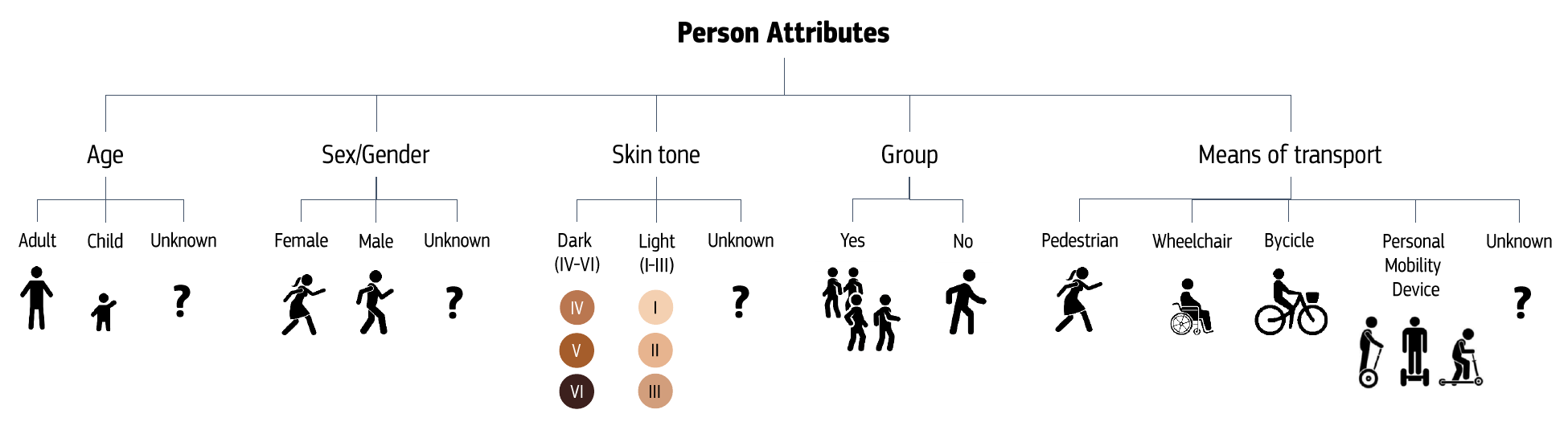}
\caption{Distribution of labelled persons attributes.}
\label{fig:personsAttributes}
\end{figure*}

\begin{figure*}[!th]
\centering
\includegraphics[width=0.8\textwidth]{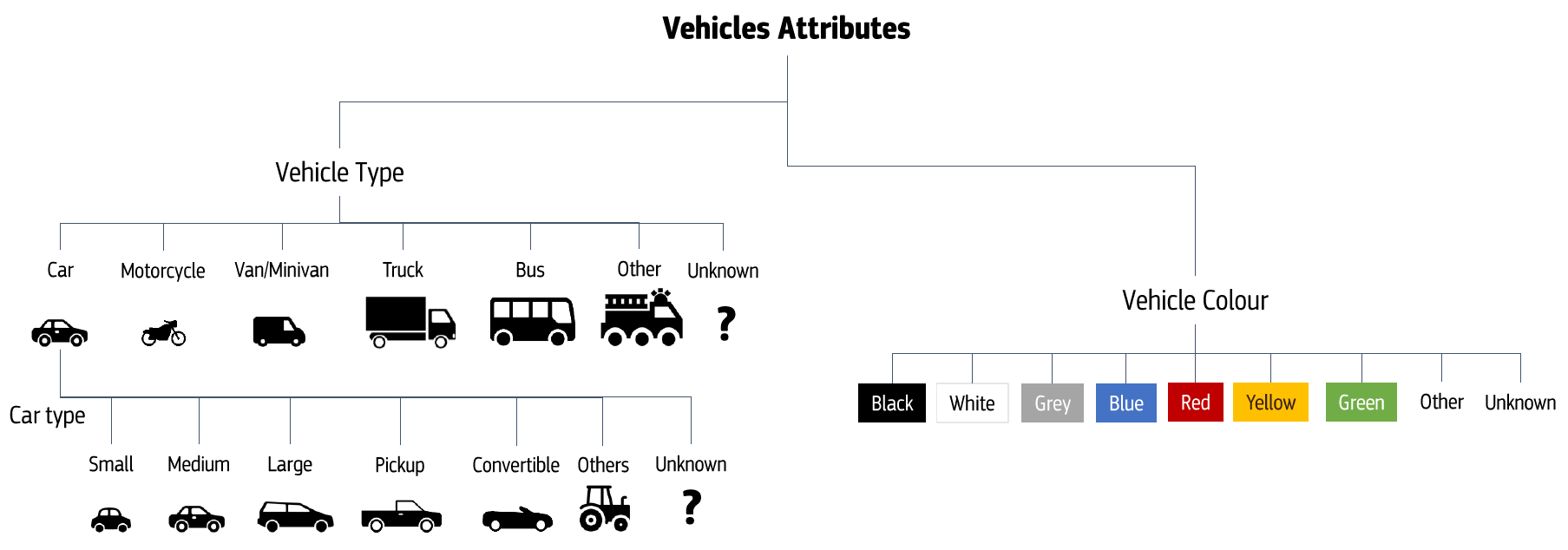}
\caption{Distribution of labelled vehicles attributes.}
\label{fig:vehiclesAttributes}
\end{figure*}

In the following, we describe the attributes considered for the two main types of agents: persons (see Fig. \ref{fig:personsAttributes}) and vehicles (see Fig. \ref{fig:vehiclesAttributes}).

\subsubsection{Persons datasets}\label{ssubsec:attributes-person}

\textbf{Age}: This attribute can have an impact on both the appearance \cite{Brandao2019} and behavior model \cite{Escobar2021} of a person-type agent. Given the typical image resolution available for pedestrian detection, bounding box size is the most important appearance factor, which is quite noticeable between children and adults. That is, for the same distance, the size of the pedestrian projected in the image is significantly smaller for children than for adults, and pedestrian detection systems are well known for reporting worse results for small agent sizes~\cite{Parra2007}. With respect to behavior, numerous studies have identified significant differences between children and adults \cite{Suzanne2003}, and in some cases, with respect to elderly people \cite{Lord2018}. However, like in \cite{Brandao2019}, only child and adult categories have been considered since identifying elderly people compared to younger adults is quite challenging for both an annotator and a perception system \cite{Wang2022}, especially at medium or low bounding box resolutions.

\textbf{Sex/gender}: Consistent with previous studies \cite{Buolamwini2018}, \cite{Brandao2019}, we assume this attribute as a binary variable according to traditional or stereotypical male/female physical appearance and morphology. The visual appearance of individuals is influenced by many factors, including both sex (e.g., sexual dimorphism, body and facial morphology~\cite{Wells2007}) and gender (e.g., clothing, hairstyle)\footnote{We acknowledge that gender is not a binary construct and that an individual's gender identity may not align with their perceived or intended gender presentation. However, for the purposes of this study, we instruct annotators to categorize based on predominantly feminine or masculine presentations even though these may not align with the individual's self-identified gender.}. Significant discrepancies in sample distribution between female and male pedestrians may affect the performance of vision-based detection systems \cite{Brandao2019}. Furthermore, there is evidence that this attribute can affect pedestrian behavior as well \cite{Holland2010}, and thus, any potential bias in the data can significantly impact the accuracy of predictive systems.

\textbf{Skin tone}: This attribute refers to skin pigmentation of person-type agents. As proposed in \cite{Wilson2019}, we broadly divide the Fitzpatrick skin type scale \cite{Fitzpatrick1975} into two groups: light skin tone corresponding to types I-III and dark skin tone corresponding to types IV-VI. Higher levels of granularity are not necessary for the purpose of the analysis and would be too complex to annotate accurately. Previous evidence on skin tone for pedestrians indicates that when individuals with dark skin tone are underrepresented in datasets, pedestrian detection systems tend to perform less accurately compared to the detection of light skin tone individuals \cite{Wilson2019}. However, this attribute is less relevant for predictive systems as, to the best of our knowledge, there is currently no evidence indicating different behaviors depending on skin tone.

\textbf{Group/individual}: Pedestrians can move either individually or in a group (i.e., two or more people moving together). This attribute can affect both detection and prediction. On one hand, when moving in a group, it is more likely to have different degrees of overlap, meaning some parts of pedestrians may not be fully visible from the camera, or the main region of a person may contain parts of other persons. These effects may affect the performance of the vision-based perception system, resulting in poorer performance in detecting people in a group compared to individuals~\cite{Zhang2018}. On the other hand, there is evidence that when pedestrians are moving or crossing as a group, they tend to behave different than when they are alone~\cite{Rasouli2020}. For example, accepting shorter gaps between vehicles to cross, or not looking at upcoming traffic. Group size exerts some form of social force over individual pedestrians~\cite{Rosenbloom2009}. It also affects pedestrian flow and speed. Therefore, action and motion prediction systems will be affected by this attribute. 

The annotation tool automatically pre-assigns agents to groups, each with a unique identifier. However, in the event of an error such as two close pedestrians crossing in opposite directions, annotators can undo the pre-selection, remove groups and re-assign agents to different groups as needed.

\textbf{Means of transport}: Appearance, motion dynamics, and decision making differ according to the means of (non-motorized) transport used by the person. This category includes pedestrians, wheelchair users, bicyclists, and users of Personal Mobility Devices (PMD) such as electric scooters, hoverboards, unicycles or segways~\cite{Laverdet2023}. Therefore, this attribute impacts both detection and prediction systems.

\subsubsection{Vehicles datasets}\label{ssubsec:attributes-vehicle}
\textbf{Vehicle type}: The appearance of vehicles is influenced by the type of vehicle, affecting its shape and size. There is evidence of bias in the detection and classification of vehicles when certain types of vehicles are underrepresented in the datasets~\cite{Corrales2021}. But the type of vehicle also influences the dynamics of movement, with behaviours that can be very diverse (for example, between a motorcycle and a bus). As depicted in Fig. \ref{fig:vehiclesAttributes}, the following categories have been considered for this attribute: car, motorcycle, van/minivan, truck, bus and others. An additional layer is used for the car category, which includes different car types and segments (more details can be seen in Table \ref{tab:guideVehicle}).

\textbf{Vehicle colour}: Apart from shape and size (vehicle type) the only attribute that influences the appearance of the vehicle for a specific pose as seen from the camera is the colour. Colour, as a feature, has been used to detect vehicles~\cite{Tsai2007}, and vehicle colour recognition is a classic problem within the field of vehicle detection and identification~\cite{Chen2014}, which shows that it is a sufficiently distinctive feature. On the other hand, although preliminary evidence suggests a correlation between vehicle colour and crash risk~\cite{Newstead2010}, this effect cannot be directly attributed to driver behavior. Rather, it is linked to the visibility of the vehicle itself and is closely related to appearance and detection. Therefore, this attribute is not considered relevant for predictive systems.

The colour labelling of vehicles is highly dependent on lighting conditions and camera characteristics, conditions that change both at the dataset level and between datasets. After several preliminary studies, it was decided to limit the number of colour categories to eight: black, white, grey, blue, red, yellow, green and others (see Fig. \ref{fig:vehiclesAttributes}). A higher level of granularity would result in larger inter-rater disagreements and labelling errors.

\subsubsection{Uncertain and conflicting cases}\label{ssubsec:conflicts}
We have provided the annotator with the category \enquote{unknown} for all attributes, except for the group, in anticipation of the possibility of uncertain or conflicting decisions. In this context, \enquote{uncertain} refers to situations in which, even with the zoom functionality, it is extremely difficult to select a specific attribute, such as the vehicle colour or skin tone during nighttime conditions. Additionally, the assignment of vehicle and car types can be challenging in certain cases, depending on the perspective and vehicle pose. This uncertainty is also present in the boundaries between categories, such as adult/child or car types. On the other hand, \enquote{conflicting decisions} refer to cases where the annotator feels that some kind of stereotypical decision is being made based on his or her own prejudices. For example, in some cases the attributes of sex/gender, age or skin tone are based on certain stereotypical characteristics that may not correspond to reality. As explained in Section \ref{subsec:method}, we allow the annotator to freely select \enquote{unknown} in such cases or make a decision based on such stereotypes.

\subsection{Annotation methodology}\label{subsec:method}
There is growing concern and awareness of widespread problems in the annotation of commonly used image benchmarks, which are subject to errors and biases \cite{Paullada2021, Northcutt2021}. Recent work highlights the importance of labelling instructions for high quality annotation \cite{Radsch2023}. 

\begin{figure}[!t]
\centering
\includegraphics[width=0.45\textwidth]{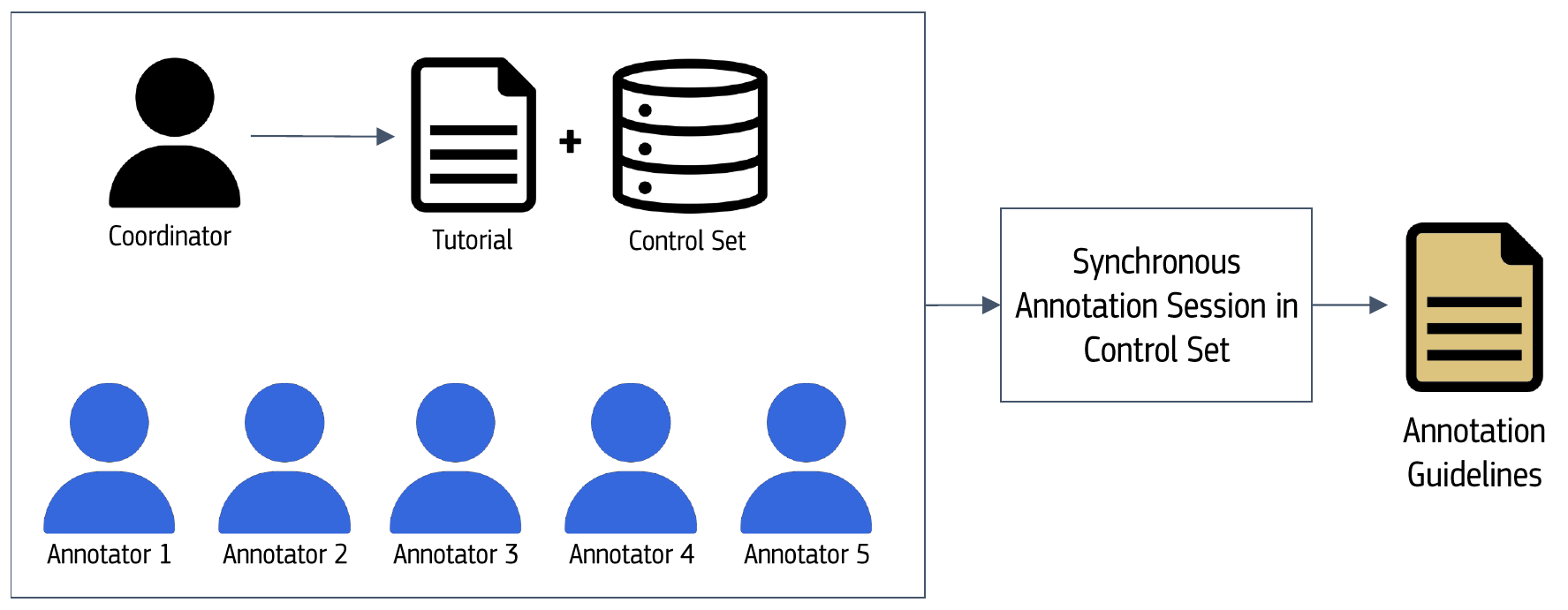}
\caption{Proposed approach to develop the annotation methodology and guidelines.}
\label{fig:guidelines}
\end{figure}

To establish a common guidelines and methodology on the different annotation attributes for all annotators, we implemented the following procedure (see Fig. \ref{fig:guidelines}). First, a basic tutorial was created for annotators to interact with the annotation tool and identify the different attributes. Next, an initial control set of annotated samples was prepared  that was as representative as possible of each dataset, while maintaining a manageable size. To define the control set, a random but representative subset of all datasets (15 images each) was selected from the available annotation tool. Based on the instructions and the control set, a synchronous labelling session was developed in which all annotators and the coordinator proceed to label the control set at the same time in the same session, on the same projected screen. For each image, each annotator manually wrote down the attributes of the agents on a piece of paper, and all results were immediately shared, compared and discussed. 

This synchronous labelling session pursued two main objectives. First, to instruct the annotators in the annotation process. At first, the differences in annotation criteria were greater, but after comparing and discussing various images, a consensus began to emerge, and the process went much faster as there were fewer and fewer differences in annotation. And second, to gather information on the consensus reached in order to incorporate it into the general annotation guidelines (see Appendices \ref{appendix:sessionguidelines}-\ref{appendix:vehicleguidelines}). All annotators should follow these guidelines to maximize the quality of the annotations, ensuring consistency between different annotators (high inter-rater agreement) and among annotations by the same annotator (high intra-rater agreement).

\section{Results}\label{sec:results}
The annotation process included to annotate a \textit{control group} of 5400 and 3000 agents for persons and vehicles respectively (the so-called inter-agreement) that was annotated by all the 5 annotators. The remaining agents were annotated only once by a single annotator. This inter-agreement was used, on a first phase, to detect biases among annotators and take corrective actions. On a second phase, it was used to perform an inter-rater analysis and evaluate the expected agreement on the annotations of the remaining agents.
\subsection{Inter-agreement results}\label{subsec:results-inter}
To measure the inter-rater agreement we selected two different metrics. 
The first one, \textbf{the percentage of disagreement}, represents the probability of an agent of being labelled not unanimously, per attribute. It is defined as the number of agents labelled differently over the total number of agents labelled.
\begin{equation}
    PD_i = \frac{1}{N\, T_i}\sum_{n=1}^{N}{T^{D}_{n,i}}
\end{equation}
where $i$ is the annotated atributes \{Age, Group, MoT, Gender, Skin\} for pedestrians and \{VehType, colour, CarType\} for vehicles, $N$ is the total number of annotators, $T^{D}_{n,i}$ is the number of agents labelled differently to any other annotator by annotator $n$ for label $i$ and $T_i$ is the total number of agents annotated for label $i$.

\begin{table*}[!ht]
\caption{Average percentage of disagreement per label and dataset}
\label{tab:PD_results}
\centering
\begin{tabular}{ccccccc}
\toprule  
    & KITTI \cite{Kitti2012} & TDC \cite{TDC-2016} & CityPersons \cite{CityPersons2017} & EuroCity Persons \cite{Eurocity2019} & BDD100K \cite{bdd100k-2020} & nuImages \cite{nuImages}\\
    \midrule
Age    & 6.17\%   & 4.01\%  & 7.86\%   & 6.99\%  & 5.65\%   & 2.88\%  \\
Group  & 11.33\%  & 3.90\%  & 5.52\%   & 6.16\%  & 3.64\%   & 4.75\%  \\
MoT    & 11.83\%  & 6.02\%  & 7.47\%   & 6.52\%  & 4.93\%   & 2.44\%  \\
Gender & 22.78\%  & 41.69\% & 17.12\%  & 29.02\% & 26.80\%  & 14.01\% \\
Skin   & 12.04\%  & 32.06\% & 17.95\%  & 28.37\% & 26.69\%  & 22.64\% \\
\bottomrule
\end{tabular}
\end{table*}

Results displayed on Table \ref{tab:PD_results} can be interpreted as the expectancy of a disagreement on labelling each one of the labels on each one of the datasets. However, those figures have been calculated in a pessimistic way, as all labels have been considered to have the same certainty, including \enquote{unclear} and \enquote{not clear}. Disagreements where some of the raters label \enquote{unclear} could have been considered to contribute less to the overall disagreement than those where opposite labels are chosen (ie. male and female). 

The second one, Fleiss' Kappa \cite{TDS_IAA}, is a statistical measure for assessing the reliability of agreement between a fixed number of raters when assigning categorical ratings to several items. It is a generalization of Scott’s Pi evaluation metric for two annotators extended to multiple annotators. Whereas Scott’s pi and Cohen’s kappa work only for two raters, Fleiss’ kappa works for any number of raters giving categorical ratings, to a fixed number of items. In addition to that, not all raters are required to annotate all items. The measure calculates the degree of agreement in classification over that which would be expected by chance. Landis and Koch \cite{Landis1977TheMO} created a table for interpreting the kappa values for a 2-annotators 2-class example, but this is not universally accepted, and it is not based on evidence but on personal opinion. The magnitude of the kappa is known to be affected by number of categories and subjects. Fleiss provided an equally arbitrary guidelines for his kappa: over 0.75 as excellent, 0.40- 0.75 as fair to good and bellow 0.40 as poor.

The formula for the computation of Fleiss' Kappa is
\begin{equation}
    k = \frac{P-P_e}{1-P_e}    
\end{equation}
where $P$ is a measure of the agreement in the annotation and $P_e$ is a measure of the balance in the labels of the annotated set. In this way, for a perfect agreement between annotators, $P$ is 1, decreasing as the agreement reduces. On the other hand, $Pe$ is 1 in a set where all the samples have the same label, and has smaller values for equally balanced sets, with a value that depends on the number of possible labels. The higher the number of possible labels the smaller the value of $P_e$. In this way, $P_e$ works as a modifier of the significance of the agreement reached with $P$ reducing the value of $k$ if the labels in the set are unbalanced.

\begin{table}[!ht]
\caption{Fleiss' Kappa per label}
\label{tab:Fleiss_results}
\centering
\begin{tabular}{cccc}
\toprule  
    & $P$ & $P_e$ & $k$ \\
    \midrule
Age    & 94.69\%	& 90.90\% & 41.67\%   \\
Group  & 94.62\%	& 39.86\% & 91.06\%   \\
MoT    & 94.96\%	& 68.78\% & 83.86\%   \\
Gender & 76.25\%	& 37.62\% & 61.92\%   \\
Skin   & 73.54\%	& 48.13\% & 48.98\%   \\
\bottomrule \toprule 
VehType   & 90.97\%	& 58.96\% & 78.00\%   \\
Colour     & 70.31\%	& 18.58\% & 63.54\%   \\
CarType   & 70.07\%	& 31.21\% & 56.49\%   \\
\bottomrule
\end{tabular}
\end{table}

The results in Table \ref{tab:Fleiss_results} indicate a good to excellent level of agreement, with $P$ ranging from $95\%$ to $70\%$ and a fair to poor balance of labels in the annotated datasets, with fair $P_e$ in gender, car colour or car type and very unbalanced labels in age and means of transport. The combination of the level of agreement and the balance in the dataset provides $k$ that are from excellent to fair being the most problematic cases the age due to a very unbalanced labeling and skin tone due to a combination of fair agreement and poor-fair balance.

The computation of Fleiss' Kappa assumes equal importance for any kind of disagreement, but, as mentioned earlier, in our sets not all the disagreements indicate the same. To give some insights about the nature of the disagreements, first we will break down some statistics about the different disagreements in Table \ref{tab:disagreement-brokendown}. 
Specifically, we will separate the disagreements by the number of raters that disagree and will remove some of the \textit{"softer"} labels such as \textit{"unknown"} or \textit{"not clear"}.
According to the number of possible labels $M$ for a given attribute and the number of raters $N$, the different outcomes are set by a combination with replacement of $N$ elements taken from a set of $M$, denoted as $C^R(M,N)$. Table \ref{tab:disagreement-brokendown} shows the number of samples for each labelling outcome split by attribute. The first row of each category shows the labels without any filtering. The second row shows the same labeling discarding the \textit{softer} disagreements.

\newcolumntype{J}{>{\raggedright\arraybackslash}m{0.1\textwidth}}
\newcolumntype{s}{>{\centering\arraybackslash}p{0.08\textwidth}}
\begin{table*}
\caption{Distribution of agreed and disagreed samples for the inter-agreement subset per category.}
\footnotesize
\centering
\scalebox{1.0}{
\begin{tabular}{Jssssssss}
\toprule
                    & &\textbf{Agreement} & \multicolumn{6}{c}{\textbf{Disagreement}}\\
Labelling outcome$^a$  & &\textbf{5}    & \textbf{4/1}  & \textbf{3/2}  & \textbf{3/1/1}    & \textbf{2/2/1}    & \textbf{2/1/1/1}  & \textbf{1/1/1/1/1} \\
$k-score$           & &1             &  0.6          & 0.4           & 0.3               & 0.2               & 0.1               & 0 \\
\midrule 
Age &   $C^R(3,5)$      & 5,319          & 336           & 169           & 10                & 4                 & -                 & -  \\
    &   $C^R(2,5)$      & 5,302          & 37            & 13            & -                 & -                 & -                 & -  \\
\midrule
Sex &   $C^R(4,5)$      & 3,505          & 1,056          & 328           & 523               & 320               & 105               & -  \\
    &   $C^R(2,5)$      & 3,433          & 149           &  35           & -                 & -                 & -                 & -  \\
\midrule
Skin tone & $C^R(4,5)$  & 3,075          & 1423          & 487           & 521               & 260               & 67                & -  \\
          & $C^R(2,5)$  & 1,047          & 352           & 203           & -                 & -                 & -                 & -  \\
\midrule
MoT &   $C^R(5,5)$      & 5,336          & 369           & 121           & 12                & 4                 & 0                 & 0  \\
    &   $C^R(4,5)$      & 5,213          & 114           & 15            & 1                 & 0                 & 0                 & -  \\
\bottomrule \toprule 
Colour &   $C^R(9,5)$    & 1,412          & 741           & 357           & 233               & 208               & 54                & 3  \\ 
      &   $C^R(7,5)$    & 1,319          & 307           & 89            & 27                & 7                 & 1                 & 0  \\ 
\midrule
Vehicle Type &   $C^R(6,5)$     & 2,483          & 277           & 187           & 33                & 16                & 12                & 0  \\ 
     &   $C^R(4,5)$     & 2,449          & 98            & 52            & 0                 & 3                 & 0                 & -  \\ 
\midrule
Car type &  $C^R(7,5)$  & 959           & 742           & 470           & 231               & 174               & 17                & 0 \\ 
         &  $C^R(5,5)$  & 842           & 393           & 246           & 18                & 9                 & 0                 & 0 \\ 
\bottomrule \\
\multicolumn{9}{l}{\footnotesize{$^a$ Combination of possible labeling groups and number of labels in each group generated in the agreement set for each category.}} \\
\multicolumn{9}{l}{\footnotesize{$C^R(n,r)$ is a combination with replacement of $r$ elements taken from a set of $n$ elements.}} \\
\end{tabular}
}
\label{tab:disagreement-brokendown}   
\end{table*}

The disagreement \enquote{level} is measured by means of the $k_{score}$ which evaluates the number of repeated labels for a single annotation. For a perfect agreement, its value is 1 and decreases with the number of different labels and its dispersion. As far as the number of possible outcomes is limited by the number of possible labels $M$ and the number of annotators $N$, the $k_{score}$ is a discrete and tabulated value that can be observed in Table \ref{tab:disagreement-brokendown}. The $k_{score}$ is computed according to eq. \ref{eq:kscore}.
\begin{equation}
\label{eq:kscore}
    k_{score} = \frac{l_1^2 + l_2^2 + \cdots + l_{m-1}^2 +  l_{m}^2 - N}{N(N-1)}
\end{equation}
where $l_i$ is the number of labels given to category $i$ by the annotators for a specific attribute and agent.



\subsubsection{Qualitative analysis}\label{subsubsec:qualitative-results}
\begin{figure*}[!t]
\centering
\subfloat{\includegraphics[width=0.25\textwidth]{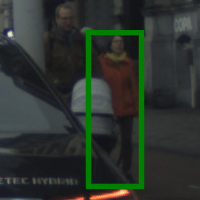}%
\label{fig_first_casea}}
\hfil
\subfloat{\includegraphics[width=0.25\textwidth]{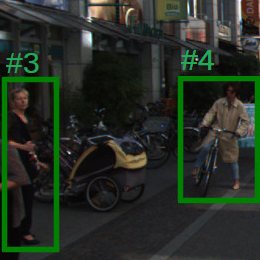}%
\label{fig_first_caseb}}
\hfil
\subfloat{\includegraphics[width=0.25\textwidth]{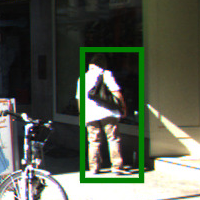}%
\label{fig_first_casec}}
\hfil
\subfloat{\includegraphics[width=0.25\textwidth]{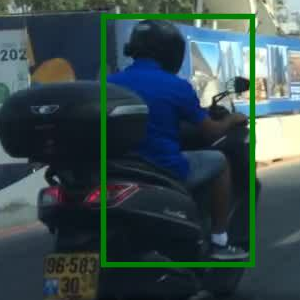}%
\label{fig_first_cased}}
\hfil
\vspace{-3.5mm}
\subfloat{\includegraphics[width=0.25\textwidth]{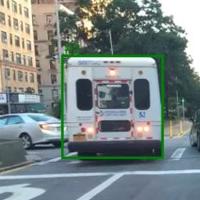}%
\label{fig_first_casee}}
\hfil
\subfloat{\includegraphics[width=0.25\textwidth]{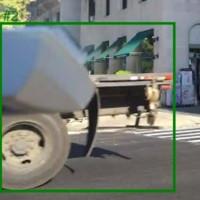}%
\label{fig_first_casef}}
\hfil
\subfloat{\includegraphics[width=0.25\textwidth]{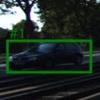}%
\label{fig_first_caseg}}
\hfil
\subfloat{\includegraphics[width=0.25\textwidth]{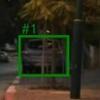}%
\label{fig_first_caseh}}
\caption{Examples of disagreements for pedestrian agents (top row) and vehicle agents (bottom row). Starting from the top left, example of "soft" disagreement on "age" \{kid, unknown, adult, kid, kid\}. Top row second picture is an example of a disagreement probably by confusion on "means of transport" \{bicycle, pedestrian,pedestrian, pedestrian, pedestrian\} for agent \#3 and \{pedestrian, bicycle, bicycle, bicycle, bicycle\} for agent \#4. Top row third picture is an example of soft disagreement in gender \{female, female, unclear, male, unclear\}. Top row last picture is an example of soft disagreement on "skin" \{light skin, dark skin, dark skin, dark skin, dark skin\}. On the bottom row first picture there is an example of "vehicle type" disagreement \{car, bus, bus, car, bus\}. On the bottom row second picture there is an example of "vehicle type" disagreement \{truck, truck, truck, unknown, truck\}. On the bottom row third picture there is an example of "soft" disagreement on "colour" \{black, unknown, black, unknown, black\}. On the bottom row last picture there is an example of "car type" disagreement \{large, small, medium, unknown, large\} }
\label{fig:example_interacion_exterior}
\end{figure*}

In addition to the numerical results, we would like to present some qualitative examples of representative disagreements on Figure \ref{fig:example_interacion_exterior}. The first one, in Figure \ref{fig:example_interacion_exterior} top row third picture, is what we consider a \enquote{soft} or \enquote{fair} disagreement in which some annotators were reluctant to assign a label to a difficult labelling situation. The picture shows a person looking at a shop window, with male morphology but wearing a bag. His/her face or hair is not visible. Three of the annotators decided that the morphology and the bag was enough to assign a gender while other two decided that it was not clear. Here we find a problem in training the annotators on the degree of certainty that they need in order to decide assigning a label. Although this was previously trained with the annotators it is very difficult to establish a solid criterion. The second one is what we could consider errors in annotation, that is usually when 4 annotators agree (4/1). In Figure \ref{fig:example_interacion_exterior} top row second picture, we have pedestrian number 3 which received \{bicycle, pedestrian, pedestrian, pedestrian, pedestrian\} and number 4 which received \{pedestrian, bicycle,  bicycle, bicycle, bicycle\} in what seems to be a clear mistake in the agent annotation for annotator 1. In agent 4 annotation of gender we have another example of \enquote{soft} disagreement as it got 3/1/1 for gender with \{female, female, unknown c, unknown nc, female\}. Examples from the vehicles dataset are equivalent in the conclusions. In Figure \ref{fig:example_interacion_exterior} bottom row last picture, we can see a vehicle that got 3 grey and 2 unknown for colour, which is considered a \enquote{soft} disagreement. On the other hand the type of the car received 2 large, one medium, one small and one unknown. The size of the car is not visible on the image, an it can only be inferred from the previous knowledge of the annotator or the appearance of the visible area. In this case, it seems a small/medium car and the two large are considered as errors in annotation.

\begin{figure}[t]
\centering
\includegraphics[width=0.49\textwidth]{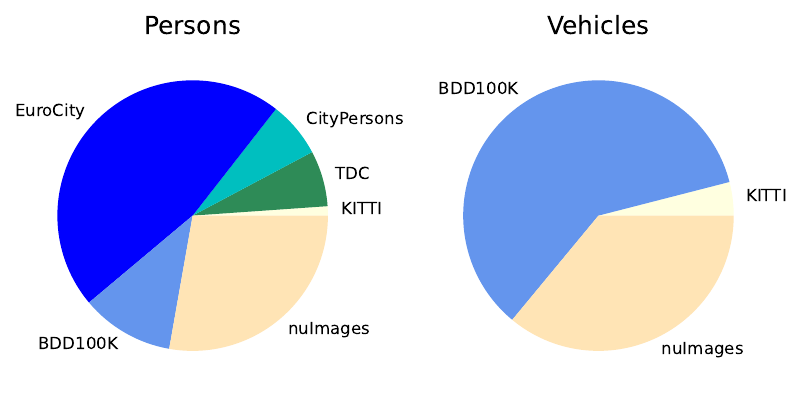}
\caption{Final distribution of annotated samples in the persons and vehicles datasets. The number of samples per dataset is more balanced than the original distribution of samples (depicted in Figure \ref{fig:pie1}).}
\label{fig:pie2}
\end{figure}

\subsection{Complete results}\label{subsec:results-complete}

\begin{figure*}[ht]
\centering
\includegraphics[width=\textwidth]{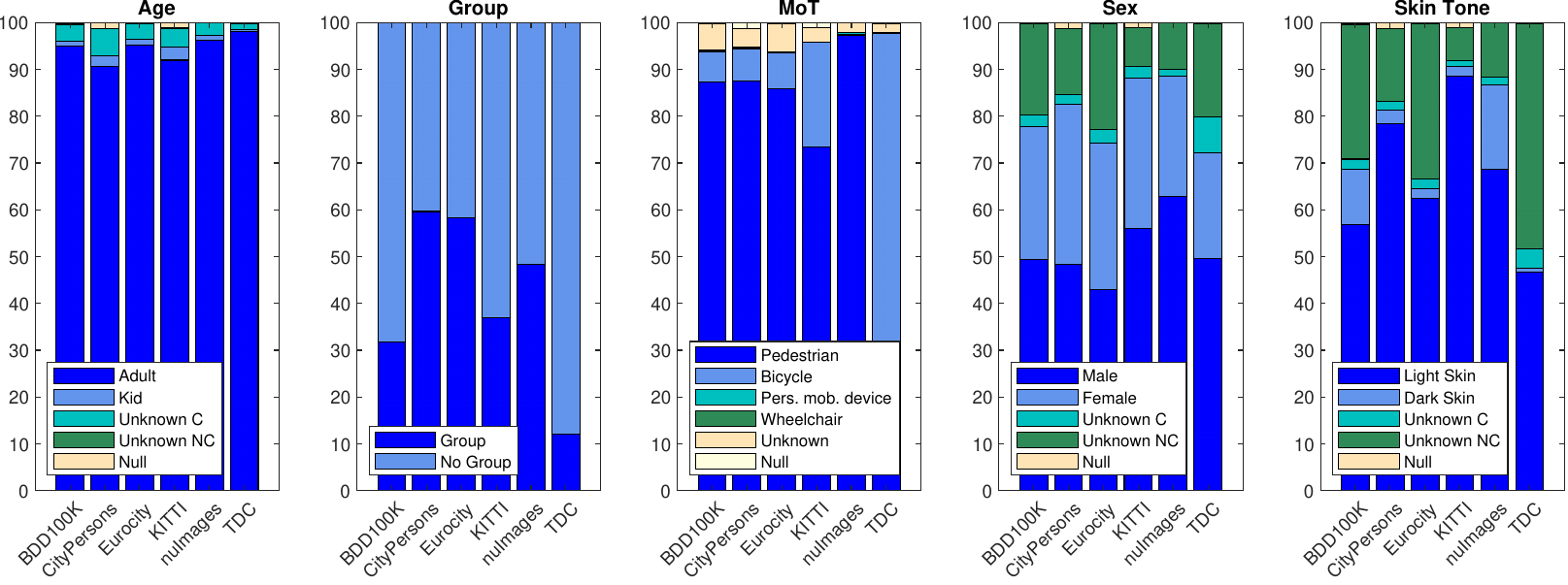}
\caption{Annotated \enquote{Age}, \enquote{Group}, \enquote{MoT}, \enquote{Sex}, and \enquote{Skin tone} distribution for the full datasets, per dataset. Diversity across \enquote{Age} and \enquote{MoT} is notably low for all datasets. \enquote{Group} diversity is reasonable in most datasets. \enquote{Sex/gender} shows a slight bias towards males in most datasets. The most diverse datasets with respect to \enquote{skin tone} are BDD100K and nuImages, although there is a considerable bias towards light skin.}
\label{fig:results_ped_stack}
\end{figure*}

\begin{figure*}
\centering
\includegraphics[width=0.78\textwidth]{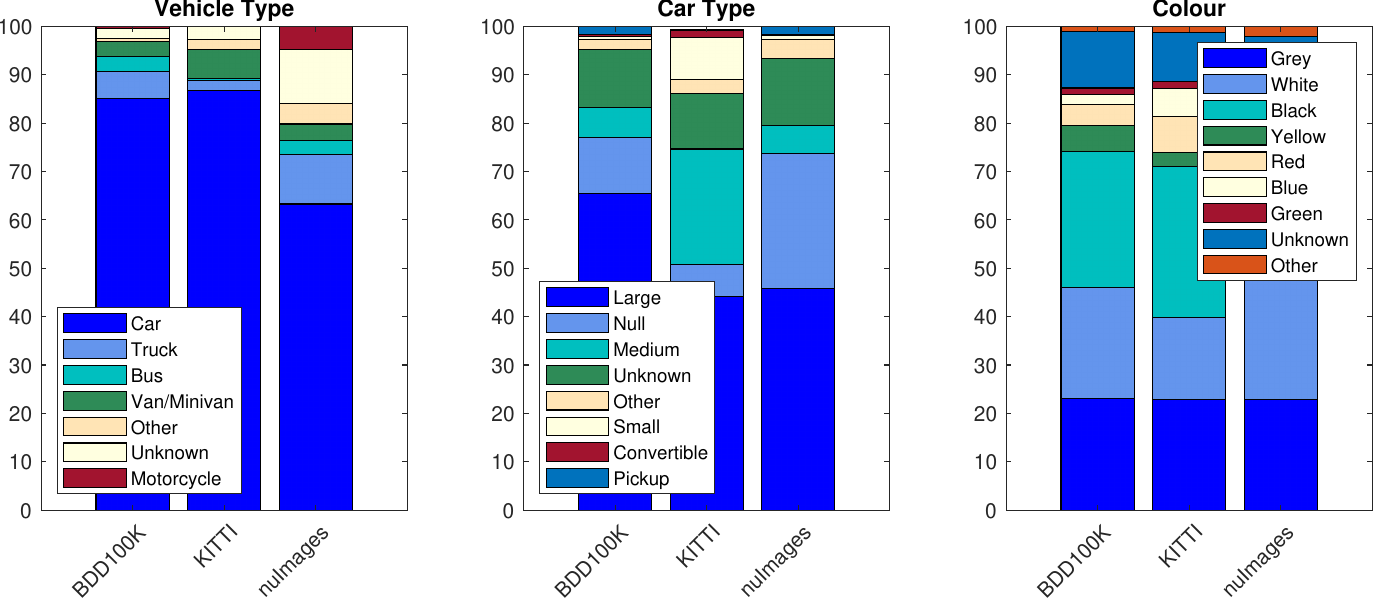}
\caption{Annotated \enquote{vehicle type}, \enquote{car type}, and \enquote{colour} distribution for the full datasets, per dataset. The most diverse dataset in terms of \enquote{vehicle type} is nuImages, although there is a considerably high bias towards cars. The distribution of \enquote{car types} largely depends on the geographical location of the datasets. The distribution of \enquote{vehicle colour} is fairly uniform across all datasets.}
\label{fig:results_veh_stack}
\end{figure*}

After analyzing the inter-agreement results, in this section, we are presenting the statistic results of the annotation of the full data-sets. The final distribution of annotated samples is depicted in Figure \ref{fig:pie2}.

Figure \ref{fig:results_ped_stack} depicts the percentages of annotation for the categories \enquote{age}, \enquote{mean of transport}, \enquote{sex}, and \enquote{skin tone} in the six datasets. Equivalently, Figure \ref{fig:results_veh_stack} depicts the percentages of annotation for the vehicle-related categories; \enquote{type of vehicle}, \enquote{type of car}, and \enquote{colour} in the three BDD100K, KITTI, and nuImages datasets.

Regarding the \enquote{age} category, and as expected from the results in section \ref{subsec:results-inter}, we can see that the datasets are extremely unbalanced, with over $90\%$ of the samples being \enquote{adult}. On average, only $1\%$ of the samples are considered kids under 14 years, and most of the differences between datasets come from the samples marked as \enquote{unknown}.


The analysis of the \enquote{mean of transport} attribute shows that most of the samples have been annotated as pedestrians across all datasets, with the exception of TDC, which is a bicycle riders dataset. KITTI has $\sim20\%$ of bicycle riders and BDD100K, CityPersons and Eurocity have $\sim7\%$. nuImages has very few samples labelled as riders. Wheelchairs and personal mobility devices are extremely underrepresented with less than $1\%$ of the samples across datasets.


Datasets are slightly biased towards with an average of $~55\%$ of the samples for \enquote{male} and $\sim35\%$ for \enquote{female} for the \enquote{sex} attribute. But it is worth noticing that the uncertainty (\enquote{unknown} labels) is high and the datasets with the lower uncertainty have the higher bias between \enquote{male} and \enquote{female}. This might indicate that if there were not as many \enquote{unknown} samples in Eurocity and TDC the bias could be higher.


In general, there is an underrepresentation of the label \enquote{dark skin} in the \enquote{skin tone} attribute, with BDD100K and nuImages presenting a higher percentage ($12\%$ and $18\%$) as both have been recorded at least partially on the USA. It is also noticeable the high \enquote{unknown} percentage in TDC ($\sim50\%$) most probably due to being a riders dataset recorded during cold weather, and thus with riders wearing many clothes.


Moving to the vehicles datasets, most of the vehicles are cars ($\sim75\%$) with higher presence of trucks in nuImages ($\sim10\%$) and BDD100K ($\sim5\%$) probably due to being recorded in the USA. Motorcycles have a low representation being a $\sim5\%$ in nuImages the higher percentage.


For the \enquote{car type} category there is a general prevalence of \enquote{large} car type, specially in BDD100K and nuImages. KITTI has a better balance between \enquote{large} and \enquote{medium}, \enquote{small}, but still large vehicles nearly doubles any other type.


The dominant labels for the \enquote{colour} attribute are black, grey, and white across the datasets. The remaining colours are also balanced, with a lower representation.

\section{Conclusions}
In this paper we presented a new set of annotations for subsets of some of the most commonly utilised visual datasets in the autonomous driving domain. To do so, we developed a specialised annotation tool that allowed multiple users to simultaneously annotate more than 90K individuals and 50K vehicles. In order to minimize common errors, biases, and discrepancies among annotators we measured different inter-agreement statistics and developed a methodology aimed at reducing  the differences in annotation for different raters. The reported inter-annotator agreement was fair to excellent according to the computed Fleiss' Kappa. Among the attributes with lowest agreement we can find gender and skin tone, probably due to the difficulty of establishing a solid criteria between the binary labelling (ie male-female) and the not clear cases. This is supported by the fact that the number of 4/1 disagreements is reduced drastically in these categories if we remove the \enquote{softer} labels such as \enquote{unclear}. We also presented a qualitative analysis of the disagreements that confirmed the above conclusions with most disagreements coming from a difficulty on establishing the limits between labels (normally in 3/2, 3/1/1, 2/2/1) and from plain errors in the annotation (most of them to 4/1).
Regarding the results of the balance for the different categories in the datasets, we found strongly under-represented labels such as \enquote{kid} with approximately a $1\%$ across datasets, with KITTI ($2.71\%$) being the highest, \enquote{dark skin} with some significant representation only in BDD100K ($11.84\%$) and nuImages ($18.01\%$), both recorded in the USA,  \enquote{personal mobility devices}, \enquote{wheelchairs}, with an average of less than $1\%$ and \enquote{buses} and \enquote{motorcycles} with a small representation in nuImages ($\sim4\%$). In general, the gender is biased towards \enquote{male} ($60-50\%$ vs $20-35\%$) but with a smaller gap in the datasets where the percentage of \enquote{unknown} is higher, which might indicate that the closer to female the higher the probability an annotator will choose \enquote{unclear}. Looking at the vehicles, \enquote{car} is clearly predominant, specially in BDD100K and KITTI with more than $85\%$ while in nuImages there is a fair amount of \enquote{truck} ($\sim10\%$) and \enquote{unknown} ($\sim11\%$) with reduces the \enquote{car} labels to $63.28\%$. Among these \enquote{car} labelled the predominant \enquote{car type} is \enquote{large} in BDD100K and nuImages, with a fair representation of \enquote{medium} and \enquote{small} only in KITTI. Finally, the distribution of vehicle colour is fairly uniform across datasets, being \enquote{black}, \enquote{white} and \enquote{grey} the most common ($\sim20\%$ each).

Current and future work focuses on evaluating pre-trained and available models for object detection and person/vehicle detection and prediction across the annotated attributes for the different datasets, to investigate potential performance differences depending on the type of agent. We will also examine the sources of biased performance and implement various bias mitigation strategies. We hope that our proposed annotation methodology and tool, as well as the publicly available annotation attributes, will contribute to the integration of fairness metrics in future evaluations of perception and prediction systems in the autonomous driving domain.


%

\appendices
\section{Annotation Session Guidelines}
\label{appendix:sessionguidelines}
To ensure the quality of the annotation please follow the indications:

\begin{itemize}
    \item Plan the annotation sessions in advance. It's better to have short, regular sessions rather than long, asynchronous ones.
    \item Set realistic, not ambitious, annotation objectives for each session, and stop once you reach them. Since the number of agents isn't always immediately accessible, consider setting time-based objectives instead.
    \item Before beginning the annotation process, prepare your table and PC. Ensure that the lighting is uniform and that you're sitting in a comfortable position.
\end{itemize}

\section{Persons Annotation Guidelines}
\label{appendix:personguidelines}
See Table \ref{tab:guidePersons}. 

\newcolumntype{G}{>{\raggedright\arraybackslash}m{0.24\textwidth}}
\newcolumntype{g}{>{\raggedright\arraybackslash}m{0.19\textwidth}}
\begin{table}[ht]
\caption{General Persons Annotation Guidelines.}
\footnotesize
\centering
\scalebox{1.0}{
\begin{tabular}{gG}
\toprule
\textbf{Criterion} & \textbf{Action} \\
\midrule 
1. Should I annotate this agent? & Yes, as long as any part is visible.\\ 
\midrule
2. I have doubts about the attributes. & Annotate as unknown if unclear. \\
\midrule
3. Are seated persons considered pedestrians? & Yes. \\
\midrule
4. What should I do with incorrectly labelled agents (poles, trees, etc.)? & All the attributes as unknown and tick \enquote{Error in agents labelling}. \\
\midrule
5. When do I add an agent to a group? & When it may affect other persons' decision to cross or not to cross. \\
\midrule
6. How do I annotate sex? & Annotate according to traditional male/female appearance and morphology. \\
\midrule
7. How do I annotate age? & Only mark \enquote{Kid} if they are clearly distinguishable (up to about 12 years). The smaller size is a must.   \\
\midrule
8. How do I annotate skin tone? & Look for exposed body parts (ankles, hands, neck, etc.) and assign by similarity with the button colour.  \\
\midrule
9. What should I do with riders on motorbikes sub-entities? & Look for exposed body parts (ankles, hands, neck, etc.) and assign by similarity with the button colour.  \\
\midrule
10. Colour in sub-entities. & Annotate the predominant colour. If unclear, annotate as unknown.  \\
\bottomrule  
\end{tabular}
}
\label{tab:guidePersons}   
\end{table}

\section{Vehicle Annotation Guidelines}
\label{appendix:vehicleguidelines}

See Table \ref{tab:guideVehicle}. 

\begin{table}[ht]
\caption{General Vehicles Annotation Guidelines.}
\footnotesize
\centering
\scalebox{1.0}{
\begin{tabular}{gG}
\toprule
\textbf{Criterion} & \textbf{Action} \\
\midrule 
1. Should I annotate this vehicle? & Yes, as long as any part is visible.\\ 
\midrule
2. I have doubts about the attributes. & Annotate as unknown if unclear. \\
\midrule
3. What should I do with incorrectly labelled agents (sub-types do not match the agent)? & Select the correct type for the agent, and \enquote{others} in the sub-type. \\
\midrule
4. Type Van/Minivan. & Designed for goods transportation. Windowless (or covered). \\
\midrule
5. Type car (Small) & A-segment cars such as: \includegraphics[width=0.24\textwidth]{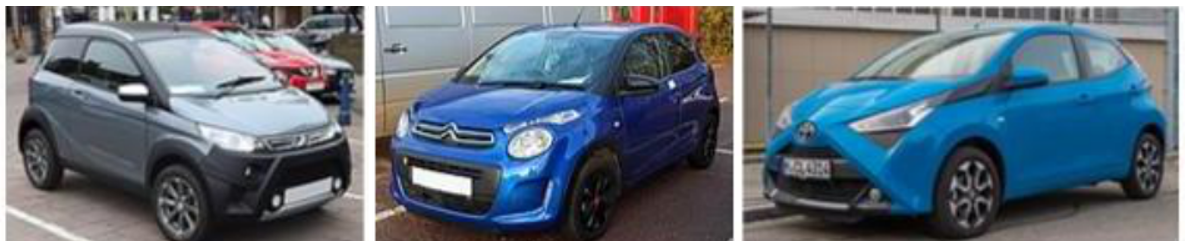} \\
\midrule
6. Type car (Medium) & B\&C-segment cars such as: \includegraphics[width=0.24\textwidth]{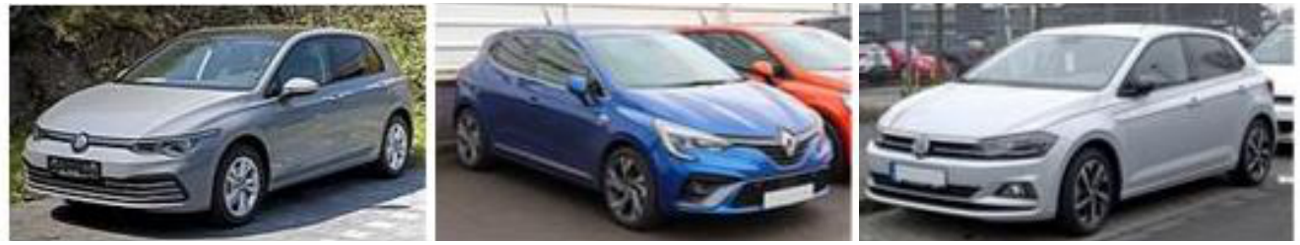} \\
\midrule
7. Type car (Large) & D\&E-segment cars such as: \includegraphics[width=0.24\textwidth]{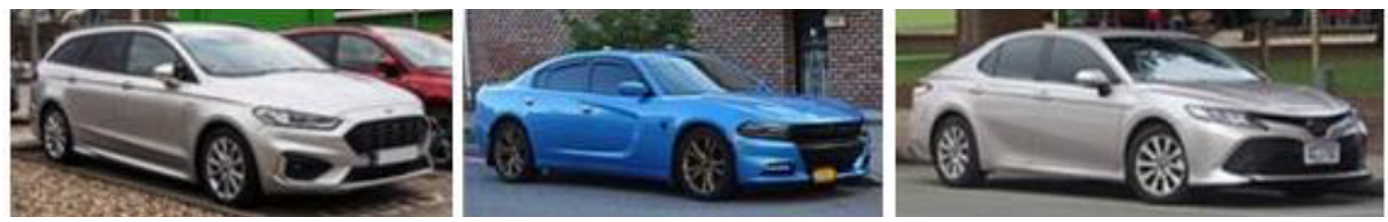} 
Also SUV 4WD, crossover, break and biggers such as:
\includegraphics[width=0.22\textwidth]{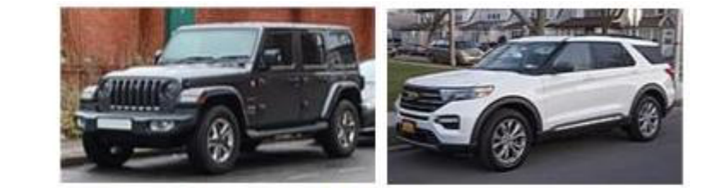}
\\
\midrule
8. Type car (Pickup) & Light-duty truck that has an enclosed cabin and an open cargo area with low sides and tailgate:  
\includegraphics[width=0.22\textwidth]{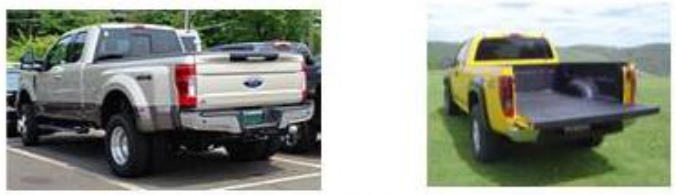}
\\
\midrule
9. Type car (Convertible) & A passenger car that can be driven with or without a roof in place:
\includegraphics[width=0.22\textwidth]{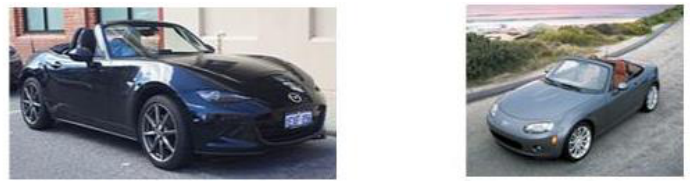}
\\
\midrule
10. Type Others & Other vehicles not fitting any other labels such as excavators or bicycles without rider.
\\
\midrule
11. How do I annotate the colour? & Annotate the predominant colour. If unclear annotate as
unknown. In nighttime without appropriate illumination, annotate unknown.
\\
\bottomrule  
\end{tabular}
}
\label{tab:guideVehicle}   
\end{table}

\section{Standardised Data Format}
\label{appendix:dataformat}
\begin{figure}[!ht]
\centering
\includegraphics[width=0.45\textwidth]{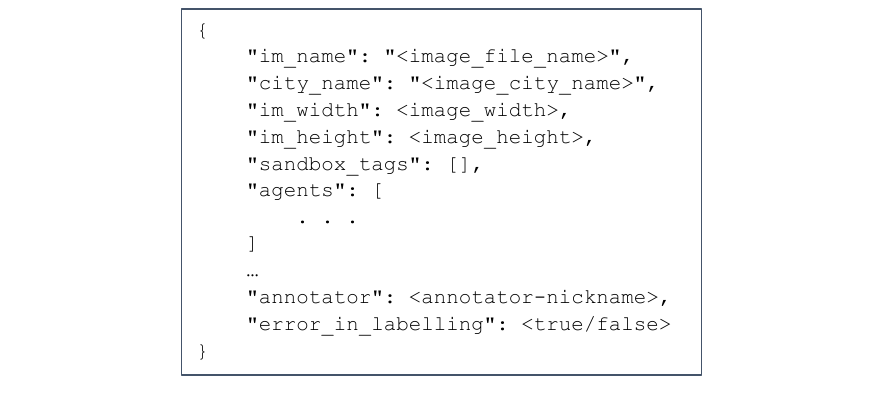}
\caption{Image metadata format.}
\label{fig:dataformatimage}
\end{figure}

\begin{figure}[!ht]
\centering
\includegraphics[width=0.45\textwidth]{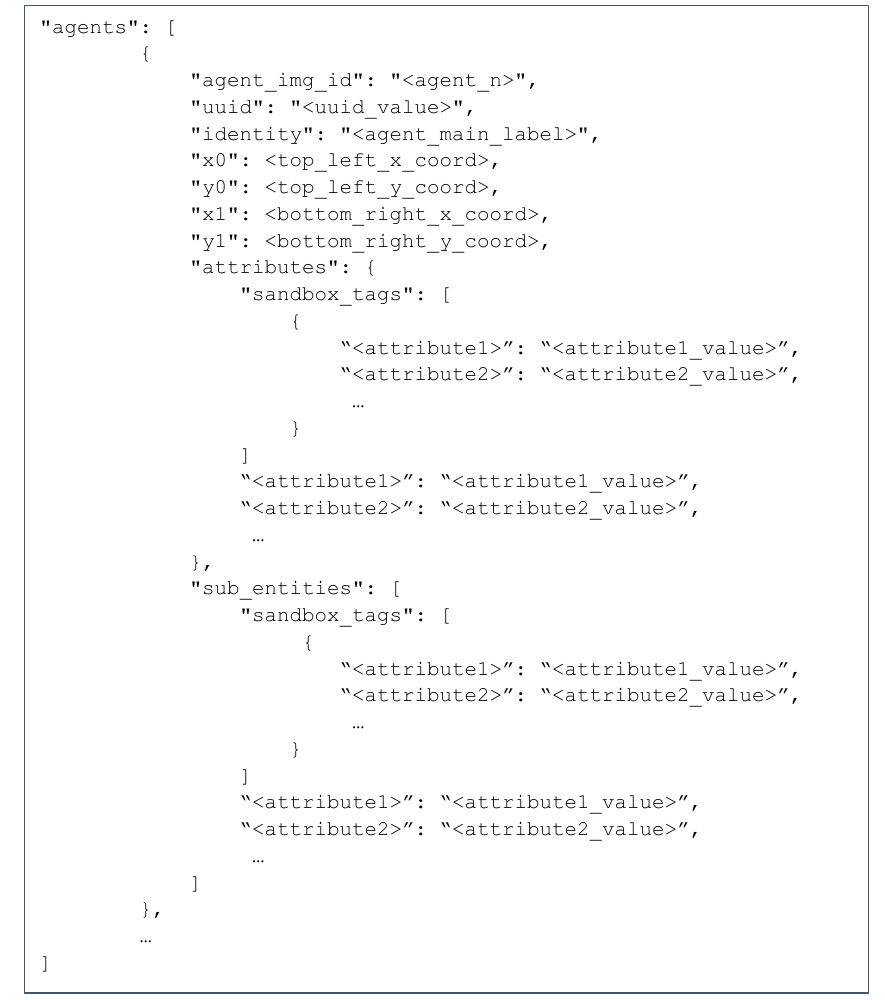}
\caption{Format of the generic attributes for each agent.}
\label{fig:dataformatagents}
\end{figure}

\begin{figure}[!ht]
\centering
\includegraphics[width=0.45\textwidth]{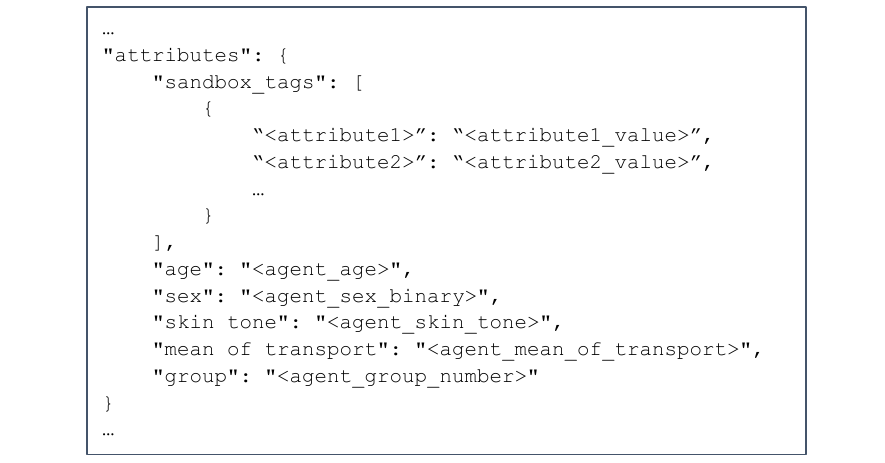}
\caption{Proposed format for the additional  annotated attributes for persons. The format for vehicle type agents is equivalent.}
\label{fig:dataformatattr}
\end{figure}

In order to keep track of every aspect of the image, the first level of the dictionary contains image metadata, including the elements listed in Fig. \ref{fig:dataformatimage}. We include an identifier or nickname of the annotator, as well as a flag to allow the image to be discarded due to an error in the original labelling of the image (all datasets contain a small number of mislabeled instances). Each agent is univocally identified using an agent image id and a unique identifier (uuid). We include the bounding box coordinates and the identity (main label). Then, we want to show all attributes for each agent, and whether they have or not any sub-entity. For the attributes, we differentiate between our annotations, and attributes which were already given by the dataset itself (sandbox tags). Sub-entities constitute agents by themselves, but depend on other agents. For example, on certain datasets we might have information about vehicles and their driver. The driver would be the main agent, and the vehicle is the sub-entity of that agent. Fig. \ref{fig:dataformatagents} depicts the format of the data for the agents. Finally, we add the new annotated attributes. An example of the attributes of the person type agent is shown in Fig. \ref{fig:dataformatattr}.

\section{Statistical Annotation Result}
\label{appendix:statisticalresults}
See Tables \ref{tab:labelling_results_ped} and \ref{tab:labelling_results_veh}.

\begin{table}[!ht]
\caption{Percentage for each of the pedestrian labels in the different datasets}
\label{tab:labelling_results_ped}
\centering
\begin{tabular}{ccccccc}
\toprule  
  \rotatebox[origin=c]{65}{Label}  & \rotatebox[origin=c]{65}{KITTI} & \rotatebox[origin=c]{65}{TDC} & \rotatebox[origin=c]{65}{CityPersons} & \rotatebox[origin=c]{65}{EuroCity} & \rotatebox[origin=c]{65}{BDD100K} & \rotatebox[origin=c]{65}{nuImages} \\
    \midrule
Adult       & 91.97   & 98.09  & 90.58   & 95.19  & 95.06   & 96.25  \\
Kid         & 2.71  & 0.32  & 2.34   & 1.32  & 0.94   & 0.95  \\
Unknown     & 4.13  & 1.40  & 5.85   & 3.20  & 3.64   & 2.72  \\
\midrule
Group       & 31.69   & 59.64  & 58.32   & 37.00  & 48.43  & 12.02  \\
No Group    & 68.31   & 40.36  & 41.68   & 63.00  & 51.57  & 87.98  \\
\midrule
Pedestrian    & 73.48   & 10.81  & 87.52   & 85.94  & 87.35   & 97.25  \\
Bicycle  & 22.35  & 86.89  & 6.94   & 7.67  & 6.49   & 0.16  \\
PMD    & 0  & 0.12  & 0.28   & 0.13  & 0.14  & 0.44  \\
Wheelchair    & 0  & 0  & 0.05   & 0.09  & 0.08   & 0.04  \\
Unknown    & 2.99  & 1.99  & 3.94 & 5.88  & 5.61 & 2.03  \\
\midrule
Male    & 56.01  & 49.65  & 48.39   & 43.05  & 49.41   & 62.86  \\
Female  & 32.22  & 28.43  & 34.23   & 31.20  & 28.43   & 25.78  \\
Unknown    & 8.23  & 19.35  & 14.10   & 22.45  & 19.35   & 9.89  \\
\midrule
Light    & 88.64   & 46.76  & 78.33   & 62.49  & 56.87   & 68.74  \\
Dark  & 2.06  & 0.68  & 3.06   & 1.94  & 11.84   & 18.01  \\
Unknown    & 6.95 & 48.12 & 15.54 & 33.02  & 28.81   & 11.48  \\
\bottomrule
\end{tabular}
\end{table}

\begin{table}[!ht]
\caption{Percentage for vehicle labels in the different datasets}
\label{tab:labelling_results_veh}
\centering
\begin{tabular}{ccccc}
\toprule  
   \rotatebox[origin=c]{65}{Category}  & \rotatebox[origin=c]{65}{Label} & \rotatebox[origin=c]{65}{KITTI} &  \rotatebox[origin=c]{65}{BDD100K} &  \rotatebox[origin=c]{65}{nuImages} \\
    \midrule
\multirow{7}{*}{\rotatebox[origin=c]{90}{Vehicle Type}} &Car    & 86.69  & 85.14  & 63.28   \\
&Truck  & 2.21  & 5.57  & 10.33 \\
&Bus  & 0.28  & 3.04  & 2.78 \\
&Van  & 6.09  & 3.19  & 3.43 \\
&Motorcycle  & 0  & 0.34  & 4.83 \\
&Unknown  & 2.78  & 2.07  & 11.08 \\
&Other    & 1.95  & 0.65  & 4.27 \\
    \midrule
\multirow{7}{*}{\rotatebox[origin=c]{90}{Car Type}} &Large    & 44.80  & 65.43  & 45.78   \\
&Medium  & 23.89  & 6.14  & 5.96 \\
&Small  & 8.61  & 0.68  & 0.7 \\
&Convertible  & 1.45  & 0.41  & 0.19 \\
&Pickup  & 0.13  & 1.66  & 1.71 \\
&Unknown  & 11.57  & 12.03  & 13.86 \\
&Other    & 2.87  & 2.07  & 3.93 \\
    \midrule
\multirow{9}{*}{\rotatebox[origin=c]{90}{Colour}} &Black    & 31.3  & 28.08  & 20.37   \\
&Grey  & 22.81  & 23.01  & 22.82 \\
&White  & 19.96  & 23.02  & 26.03 \\
&Red  & 7.48  & 4.18  & 5.37 \\
&Blue  & 5.84 & 2.08  & 4.44 \\
&Yellow & 2.77  & 5.50  & 2.75 \\
&Green & 1.47  & 1.39  & 0.56 \\
&Unknown  & 10.09  & 11.60  & 15.49 \\
&Other & 1.30  & 1.16  & 2.17 \\
\bottomrule
\end{tabular}
\end{table}

\section*{Disclaimer}
The views expressed in this article are purely those of the authors and may not, under any circumstances, be regarded as an official position of the European Commission. 

\ifCLASSOPTIONcompsoc
  \section*{Acknowledgments}
\else
  \section*{Acknowledgment}
\fi

This work was mainly funded by the HUMAINT project by the Directorate-General Joint Research Centre of the European Commission. It was also partially funded by Research Grants PID2020-114924RB-I00 and PDC2021-121324-I00 (Spanish Ministry of Science and Innovation).

\ifCLASSOPTIONcaptionsoff
  \newpage
\fi



%

\bibliographystyle{IEEEtran}
\bibliography{IEEEabrv,references}

\end{document}